\documentclass{article}

\PassOptionsToPackage{numbers, compress}{natbib}
\usepackage[preprint]{neurips_2024}

\usepackage[utf8]{inputenc} % allow utf-8 input
\usepackage[T1]{fontenc}    % use 8-bit T1 fonts
\usepackage{hyperref}       % hyperlinks
\usepackage{url}            % simple URL typesetting
\usepackage{booktabs}       % professional-quality tables
\usepackage{amsfonts}       % blackboard math symbols
\usepackage{nicefrac}       % compact symbols for 1/2, etc.
\usepackage{microtype}      % microtypography
\usepackage{xcolor}         % colors
\usepackage{graphicx}
\usepackage{amsmath}
\usepackage{subcaption}

\title{Differential Adjusted Parity for Learning Fair Representations}

\author{%
  Bucher Sahyouni\\
  Centre for Vision, Speech and Signal Processing\\
  University of Surrey\\
  Guildford, Surrey GU2 7XH\\
  \texttt{bs00826@surrey.ac.uk} \\
  \And
  Matthew Vowels \\
  University of Lusanne \\
  CH-1015 Lausanne  \\
  \texttt{matthew.vowels@unil.ch} \\
  \AND
  Liqun Chen \\
  Unviersity of Surrey \\
  Guildford, Surrey GU2 7XH \\
  \texttt{liqun.chen@surrey.ac.uk} \\
  \And
  Simon Hadfield \\
  Centre for Vision, Speech and Signal Processing\\
  University of Surrey\\
  Guildford, Surrey GU2 7XH \\
  \texttt{s.hadfield@surrey.ac.uk}\\
}

\begin{document}

\maketitle

\begin{abstract}
The development of fair and unbiased machine learning models remains an ongoing objective for researchers in the field of artificial intelligence. We introduce the Differential Adjusted Parity (DAP) loss to produce unbiased informative representations. It utilises a differentiable variant of the adjusted parity metric to create a unified objective function. By combining downstream task classification accuracy and its inconsistency across sensitive feature domains, it provides a single tool to increase performance and mitigate bias. A key element in this approach is the use of soft balanced accuracies. In contrast to previous non-adversarial approaches, DAP does not suffer a degeneracy where the metric is satisfied by performing equally poorly across all sensitive domains. It outperforms several adversarial models on downstream task accuracy and fairness in our analysis. Specifically, it improves the demographic parity, equalized odds and sensitive feature accuracy by as much as 22.5\%, 44.1\% and 40.1\%, respectively, when compared to the best performing adversarial approaches on these metrics. Overall, the DAP loss and its associated metric can play a significant role in creating more fair machine learning models.
\end{abstract}

\section{Introduction} \noindent As artificial intelligence (AI) and machine learning (ML) models become increasingly prevalent, the need for responsible, fair, and unbiased machine learning is critical \cite{barocas2016}. Many machine learning models heavily rely on learned representations which distill complex data into compressed forms, capturing key patterns for efficient learning and prediction \cite{bengio2013}. However, they may also encode sensitive attributes like gender or race, thus leading to biased decisions that can perpetuate societal inequities. 

The recognition of these biases has spurred the development of debiasing techniques. These methods aim to remove sensitive information while maintaining task performance, a balance that is challenging to achieve. Among various approaches, adversarial training has become popular, introducing an adversarial model during training to learn sensitive features. This process, often a min-max game, where the main model competes with the adversarial model making sensitive features more difficult to learn while focusing on the main task \cite{zhang2018}, is conceptually appealing but unstable and computationally intensive, requiring careful hyperparameter tuning and rigorous experimentation \cite{arjovsky2017}.

In response to these challenges, this paper introduces a new approach to debiasing. We propose a differentiable variant of the adjusted parity metric as a single objective function that considers both task classification accuracy and its invariance across sensitive feature domains, without an adversarial componenet. This approach simplifies the learning process by focusing on cooperative elements of the loss that improve performance across protected characteristics. Our technique extends to multi-class problems, ensuring fairness across various sensitive features. Our approach aims to simplify the debiasing process and enhance the stability during training, contributing to the ongoing efforts to foster ethical AI \cite{crawford2016}.

The rest of the paper is structured as follows: subsequent sections cover popular debiasing techniques, the adjusted parity metric, our experimental methodology, and a discussion of the results, concluding with insights into potential future work.

\section{Background and Related Work}
\label{sec:background}
Below we will first discuss modern adversarial techniques for debiasing. We will then cover the non-adversarial techniques which are more closely related to this paper. This is followed by a formalisation of the various definitions of fairness found in the literature.

\subsection{Adversarial Debiasing}

Adversarially Learned Fair Representations, ALFR (Edwards and Storkey, 2016), is one of the earliest models developed to reduce bias and learn fair informative representations. It utilises an adversary which tries to predict the sensitive feature from the representations. The autoencoder tries to make these predictions as difficult as possible. It employs task and reconstruction losses to ensure the representation are informative for the downstream task, and an additional sensitive feature loss to remove sensitive feature related information.

The Conditional Fair Representations, CFair, approach \cite{zhao2020} stands as a seminal method aiming for accuracy parity. Within the confines of conventional fair adversarial networks, CFair augments the original adversarial constraints and adopts conditional error constraints. 
%The objective function for \textit{CFair} is expressed as:
%\begin{align}
% \mathcal{L} = &\min_{h, g} \max_{h', h''} \text{BER}_{\mathcal{D}}(h(g(X)) || Y) \notag \\
% &- \lambda \left( \text{BER}_{\mathcal{D}_0}(h'(g(X)) || S) + \text{BER}_{\mathcal{D}_1}(h''(g(X)) || S) \right)
% \end{align}
%where \(\text{BER}\) signifies the Balanced Error Rate (Menon and Williamson, 2018), while \(\mathcal{D}_k\) represents instances associated with the sensitive attribute \(k\).

Learning Adversarially Fair and Transferable Representation, LAFTR \cite{madras2018}, is similar to CFAIR but it uses one adversary instead of two and an L1 instead of a cross entropy loss to debias the representations. It still utilises a global cross-entropy loss for the target variable.

\subsection{Non-adversarial Debiasing}
Outside of adversarial learning, Fairness by Compression (FBC) \cite{gitiaux2021} advocates for the use of binary compression to mitigate sensitive elements in representations. They establish that the cross-entropy between \( P(z) \) and \( Q \) stands as the upper bound for the entropy \( H(z) \). In this context, \( P(z) \) delineates the distribution of a factorized representation, and \( Q \) is utilized to predict \( z_i \) based on \( \{ z_0, z_1, \dots , z_{i-1} \} \). The FRC model \cite{quan2022} aims to mitigate the influence of sensitive factors in the data representation by adjusting the correlation between the representation and the sensitive vectors.
% Their proposed objective function is
% \vspace{-0.1cm}\begin{equation}
% \mathcal{L} = \min_{z} \mathop{\mathbb{E}}_{x,z,a}[-\log(P(x|z, a))] + \beta \textit{CE}(P, Q)
% \vspace{-0.2cm}\end{equation}
%where \( \textit{CE} \) signifies the cross-entropy function, and \( \beta \) modulates the compression rate.

% The objective function for FRC is formulated as
% \begin{align}
%     \mathcal{L} = &\mathop{\mathbb{E}}_{q(Z|X)}[\log p(X|Z)] - \beta \text{DKL}(q(Z|X) || p(Z))\notag \\ 
%     & + \gamma (\text{Corr}(S, Z_P) - \text{Corr}(S, Z_N))
% \end{align}
%where \(Z_P\) and \(Z_N\) denote the representation vectors that are respectively associated and dissociated with sensitive information.

BFA \cite{quan2023} draws inspiration from the correlation coefficient constraints used in FRC. The primary ambition is to minimize the correlation between sensitive information and prediction error (as opposed to minimising correlation between sensitive information and representations in FRC), aiming to maintain high predictive accuracy.

The Variational Fair Autoencoder, VFAE \cite{louizos2016}, comprises of a variational autoencoder (VAE) instead of an autoencoder and employs an addititional maximum mean discrepancy (MMD) loss to ensure less sensitive information, which may be correlated to the target task, leaks into the learned representation. MMD minimises the mismatch in moments between the marginal posterior distributions for the different sensitive features.

These techniques exhibit greatly improved training stability compared to the adversarial training approaches. However, they struggle to achieve good task performance (i.e. developing useful representations). It can be observed that many of these techniques experience a degeneracy where the loss function can be satisfied by performing equally poorly across all sensitive domains. In contrast our proposed approach maintains the benefits of non-adversarial training, while removing this degeneracy.

\subsection{Measures of Fairness}
Fairness in machine learning has been extensively studied, and there exist a range of metrics which measure different aspects of it. Our primary contribution is the introduction of a new differentiable fairness metric, thus necessitating a brief overview of commonly used metrics in our model evaluation. %Thus we will first briefly formalize the metrics that are commonly used in the field and that we will use in the evaluation of our model.
%These metrics include demographic parity, equalized odds, equal opportunity, and disparate impact.

Demographic parity, also known as statistical parity or group fairness, requires that the selection rate (the rate at which individuals are positively classified) should be the same across all demographic groups. Mathematically, if \( Y \) is the predicted label and \( A \) denotes the demographic group, demographic parity is defined as:
\vspace{-0.1cm}\begin{equation}
P(Y=1|A=0) = P(Y=1|A=1).
\end{equation}
This implies that the algorithm should be independent of the sensitive attribute \( A \) \cite{dwork2012}, which can be limiting if the attribute is relevant to the outcome \cite{hardt2016equality}.% However, demographic parity can be limiting in situations where the sensitive attribute is indeed relevant to the outcome. This and other flaws are discussed in detail in the seminal work \cite{hardt2016equality}.

The equalized odds fairness metric demands true positive rates and false positive rates to be equal across demographic groups. Mathematically, if $\hat{Y}$ is the true label:
\vspace{-0.1cm}\begin{equation}
P(Y\!=\!1|\hat{Y}\!=\!1, A\!=\!0) \!=\! P(Y\!=\!1|\hat{Y}\!=\!1, A\!=\!1)
\end{equation}
and
\vspace{-0.1cm}\begin{equation}
P(Y\!=\!1|\hat{Y}\!=\!0, A\!=\!0) \!=\! P(Y\!=\!1|\hat{Y}\!=\!0, A\!=\!1).
\end{equation}
Equalized odds aims for outcome independence across demographic groups when conditioned on the true label \cite{hardt2016}. %This means the classifier’s outcomes, both in terms of correctly identified positive cases and incorrectly identified negative cases, should be independent of the demographic group when conditioned on the true label \cite{hardt2016}. However, maintaining equalized odds can often be challenging in the presence of significant imbalances in the underlying data distributions among different groups.

%The equal opportunity metric, a subset of equalized odds, solely focuses on the true positive rates. In essence, equal opportunity requires the algorithm to have similar true positive rates across different demographic groups. This is mathematically represented as:
%
%\vspace{-0.1cm}\begin{equation}
%P(Y\!=\!1|\hat{Y}\!=\!1, A\!=\!0) \!=\! P(Y\!=\!1|\hat{Y}\!=\!1, A\!=\!1).
%\vspace{-0.2cm}\end{equation}
%
%Equal opportunity thus ensures that the proportion of correctly classified positive outcomes is the same across all demographic groups \cite{barocas2016}.

%Disparate impact refers to a scenario where a decision, rule, or policy has a discriminatory effect on a protected class, even though the explicit rules appear to be neutral. In terms of machine learning, disparate impact can be quantified as the ratio of the probabilities of positive outcomes in the protected group to that in the non-protected group. Thus, disparate impact is defined as:
%
%\vspace{-0.1cm}\begin{equation}
%\frac{P(Y=1|A=1)}{P(Y=1|A=0)}.
%\vspace{-0.2cm}\end{equation}
%
%A value of 1 indicates no disparate impact, and U.S. legal guidelines suggest that a value below 0.8 may be indicative of significant disparate impact \cite{feldman2015}.

These metrics, while essential for assessing fairness, have inherent trade-offs and limitations. It is generally impossible to satisfy all these conditions simultaneously when the base rates differ across groups \cite{chouldechova2017}. They focus on ensuring parity in predictions without necessarily considering the impact on overall task accuracy \cite{dwork2012,hardt2016equality}.

%While these metrics provide valuable tools for assessing fairness, they are not panaceas. There are trade-offs and tensions among them, and it is generally impossible to satisfy all these conditions simultaneously when the base rates differ across groups \cite{chouldechova2017}. Thus, the selection of appropriate fairness metrics depends heavily on the specific context and ethical considerations.

%When assessing the fairness of machine learning models, these metrics primarily focus on ensuring parity in prediction outcomes across different demographic groups. These metrics do not explicitly take into consideration the downstream task's classification accuracy \cite{dwork2012,hardt2016equality}.

%Specifically, demographic parity and disparate impact concentrate on the balanced positive prediction rates across various groups, disregarding the actual labels of the instances. Similarly, equalized odds and equal opportunity are concerned with similar true positive rates or error rates across groups but do not explicitly account for overall accuracy. Therefore, while these metrics ensure that models do not disproportionately disadvantage any particular group, they may do so at the cost of reducing the model's overall predictive performance \cite{corbettdavies2018}.

Recognizing the need to balance accuracy and fairness, there's a growing emphasis on metrics that integrate classification performance, creating more robust and equitable machine learning systems suitable for real-world applications \cite{grgichlaca2018}. This unified approach fosters a comprehensive evaluation and comparison of debiasing models, ensuring effectiveness in predictions while maintaining fairness.

%In the real world, accuracy and fairness are both critically important. Decisions based on machine learning models often have significant consequences, such as granting loans, admitting students to universities, or diagnosing diseases. In these situations, sacrificing accuracy can lead to suboptimal outcomes, while disregarding fairness can perpetuate systemic bias.

%Thus, there is a growing need for a unified metric that balances both classification accuracy and fairness. This approach allows for a more comprehensive comparison and evaluation of debiasing models, ensuring not just fairness but also effectiveness in prediction. By incorporating classification accuracy into fairness metrics, we can build more robust and equitable machine learning systems that are better suited for real-world applications \cite{grgichlaca2018}.

\section{The Adjusted Parity Metric}
\label{sec:AP}
As an initial step towards solving these issues, \cite{vowels2020} introduced a parity metric for evaluating domain invariance. This metric accommodates both discrepancies in accuracy across domains and normalised classifier or regressor performance to provide a single unified value for comparing debiasing models.

The adjusted parity metric was originally expressed for binary domains as:
\vspace{-0.1cm}\begin{equation}
    \Delta_{\text{adj}} = \bar{S} (1 - 2\sigma),
\end{equation}
where \( \sigma \) represents the standard deviation of the normalised classifier accuracy across the domains,  and\( \bar{S} \) denotes the average accuracy over the domains.

We extend this definition to an arbitrary number of domains:
\vspace{-0.1cm}\begin{equation}
    \Delta_{\text{adj}} = \frac{\bar{S}-S^R}{1-S^R}\left(1 - \frac{\sigma}{\gamma}\right),
\label{eq:diffpar2}
\end{equation}
where $S^R$ is the baseline accuracy of a random predictor, and $\gamma$ is the maximum standard deviation across domains. This serves to normalise the metric between [0,1].
%To illustrate, consider a scenario where there's an equal chance of predicting any of the 10 digits in the MNIST dataset by random chance. Here, the baseline would be 0.1.

The introduction of $\gamma$ in this paper extends the metric to sensitive characteristics with more than two domains. For any even number of domains ($N$) the value remains at $\gamma=0.5$ as in the original formulation. However for an odd number of domains: 
\vspace{-0.1cm}\begin{equation}
    \gamma = \sqrt{\frac{1}{4} \left(1 - \frac{1}{N^2}\right)}.
    \label{eq:gamma}
\end{equation}
Please see the appendix for a full derivation.

The implications of this metric are twofold. Firstly, any classifier that exhibits either minimal consistency or accuracy will yield \(\Delta_{\text{adj}} = 0\). Conversely, only a classifier that demonstrates maximal consistency and accuracy will result in \(\Delta_{\text{adj}} = 1\). Figure \ref{fig6} in Appendix shows how \(\Delta_{\text{adj}}\) changes with $\overline{S}$ for various $\sigma$. The motivation behind developing this metric stems from an essential understanding that invariance to a domain or attribute does not necessarily equate to reliable classification. A representation must also be informative for the intended task to be deemed effective.

\section{Differential Adjusted Parity}
\label{sec:DAP}
%\subsection{Soft Accuracies}
In order to propose a differentiable variant of the adjusted parity metric from equation \ref{eq:diffpar2}, we must first rely on a differentiable variant of the accuracy measure $S$.
%The conventional accuracy of a classifier is a widely used evaluation metric, but it suffers from two major issues: it is non-differentiable, and it can yield misleading results when the classes are imbalanced. This necessitates the use 
In this paper, we propose the use of  ``Soft Balanced Accuracy''. This is a differentiable form of balanced accuracy, which allows us to simultaneously deal with the challenges common in imbalanced problems.

% Let's consider a binary classification task where the output probability vector from a classifier is of size two, denoting the probabilities of the two classes. We denote these probabilities as $\text{probs} = [p_0, p_1]$ where $p_0$ and $p_1$ are the predicted probabilities of the first and second class respectively. Let's also assume that the true labels for the instances are represented as $\text{labels}$, where $\text{labels}[i]$ is 1 if instance $i$ belongs to the second class and 0 otherwise.

% In conventional accuracy calculation, we apply an argmax operation on $\text{probs}$ or by thresholding them to get the predicted class and compare it with $\text{labels}$. However, this operation is non-differentiable, i.e., it does not allow backpropagation of gradients during training.

% Instead, we can compute a soft version of accuracy that uses the probabilities directly without needing to apply the argmax function. For instance, the Soft True Positive (TP) for class 1 can be computed as the sum of the predicted probabilities for instances that truly belong to class 1, or mathematically: $\text{TP\_soft} = \sum((\text{labels} == 1) \times p_1)$. Similarly, for class 0, $\text{TP\_soft} = \sum((\text{labels} == 0) \times p_0)$. This directly corresponds to correctly predicting instances of each class.

The soft accuracy is computed by omitting the ``argmax'' function from a standard classification accuracy metric. In other words, for a vector of predicted class probabilities $P(x)$ given input $x$ and a one-hot encoded label vector $L_x$, the vector of soft True Positive rates for all classes is:
\vspace{-0.1cm}\begin{equation}
\text{TP} = \sum_x P(x) \odot L_x,
\end{equation}
where $\odot$ represents the Hadamard product.

Analogously, the vector of Soft False Positives (FP) for each class would be computed as the sum of the predicted probabilities for instances that are incorrectly predicted as that. Given the inverted (one-cold) label vector $\bar{L}_x$:
\vspace{-0.1cm}\begin{equation}
\text{FP} = \sum_x P(x) \odot \bar{L}_x.
\end{equation}
% : $\text{FP\_soft} = \sum((\text{labels} == 0) \times p_1)$. Similarly, for class 0, $\text{FP\_soft} = \sum((\text{labels} == 1) \times p_0)$.

We can similarly compute the vectors of Soft True Negatives (TN) and Soft False Negatives (FN) as
\vspace{-0.1cm}\begin{equation}
\text{TN} = \sum_x (1-P(x)) \odot \bar{L}_x,
\vspace{-0.2cm}\end{equation}
\vspace{-0.1cm}\begin{equation}
\text{FN} = \sum_x (1-P(x)) \odot L_x.
\end{equation}
%
%using the probabilities. For class 1, $\text{TN\_soft} = \sum((\text{labels} == 0) \times (1 - p_1))$ and $\text{FN\_soft} = \sum((\text{labels} == 1) \times (1 - p_1))$. Similarly, for class 0, $\text{TN\_soft} = \sum((\text{labels} == 1) \times (1 - p_0))$ and $\text{FN\_soft} = \sum((\text{labels} == 0) \times (1 - p_0))$.
%
Although it may be obvious, it is worth pointing out that these soft variants of the TP, TN, FP, and FN are easily differentiable, as they are computed directly from the predicted probabilities. It is also worth pointing out that the classes referred to here are based on the output task, and differ from the sensitive characteristic domains of equation \ref{eq:gamma}.
%This allows us to backpropagate gradients during training, enabling the model to learn from its errors.

Given the above, we could compute the per-class accuracy vector as
%Now, accuracy for a single class can be computed as 
\vspace{-0.1cm}\begin{equation}
\text{Acc} = \frac{(\text{TP} + \text{TN}) }{ (\text{TP} + \text{TN} + \text{FP} + \text{FN})}.
\end{equation}
However, this measure can be misleading when classes are imbalanced. For instance, in a dataset where a single class represents 90\% of the data, a naive classifier that always predicts the dominant class will have an accuracy of 90\%.

To address this, we calculate the Soft Balanced Accuracy, which is the average of the per-class recall. In other words
\vspace{-0.1cm}\begin{equation}
S = \frac{1}{C} \left|\frac{\text{TP} }{ (\text{TP} + \text{FN})}\right|^1,
\end{equation}
where $C$ is the number of classes and $|.|^1$ represents the L1 norm.
%Recall for a class is computed as $\text{TP} / (\text{TP} + \text{FN})$, and hence, $\text{Balanced Accuracy} = 0.5 \times (\text{Recall\_class0} + \text{Recall\_class1})$. In the case of Soft Balanced Accuracy, it becomes $0.5 \times ((\text{TP\_soft0} / (\text{TP\_soft0} + \text{FN\_soft0})) + (\text{TP\_soft1} / (\text{TP\_soft1} + \text{FN\_soft1})))$. 
This gives equal weighting to all classes irrespective of their prevalence in the dataset.%, thereby offering a more honest measure of the classifier's performance.

%Soft Balanced Accuracy provides a differentiable metric for evaluating and training classifiers, especially when dealing with imbalanced data. It allows for gradient-based optimization and provides an incentive for the model to perform equally well on all classes, enhancing the overall fairness of the model.

To return to the definition of $\Delta_{adj}$ in Section \ref{sec:AP}: we propose computing this soft balanced accuracy $S$ 
independently on subsets of the dataset corresponding to each sensitive domain. The mean and standard deviation of $S$ across these domains can then be substituted for $\bar{S}$ and $\sigma$ in equation \ref{eq:diffpar2} (once again noting that the set of task labels is not the same as the set of sensitive characteristics). By substituting the balanced soft accuracy into the adjusted parity metric, we can obtain a differential adjusted parity (DAP) loss which we can use to train a model. This loss encourages improvements in task accuracy across all labels, weighted by their prevalence and the current relative task performance, with no adversarial component.
%to improve the average soft balanced accuracy and reduce the standard deviation of the soft balanced accuracies across domains, thus improving task performance, while minimising the inconsistency and differences for different sensitive feature groups. 
Intuitively, minimising task prediction inconsistency across sensitive domains would minimise the mutual information between the representations and the sensitive feature. This leads to less information regarding the sensitive feature being encoded in the representations and thus better demographic parity and equalised odds scores.

\section{Evaluation protocol}
\label{sec:experiments}

In our experiments, we've chosen to focus on two widely-acknowledged datasets in fairness research: the Adult dataset and the COMPAS dataset. In both cases we train a network using a combination of our DAP metric and the standard task cross-entropy loss ($\mathcal{L}_{ce}$). We also introduce weighting hyperparameters $\beta$ and $\Omega$ which control the contribution of standard deviation term and $\mathcal{L}_{ce}$ respectively.

\subsection{Adult Dataset}

The Adult dataset, often referred to as the ``Census Income'' dataset, originates from the UCI Machine Learning Repository \cite{dua2017}. It comprises demographic data extracted from the 1994 Census Bureau database. The primary task for this dataset is binary classification: predicting whether an individual earns more than \$50k annually based on attributes like age, occupation, education, and marital status.

One notable characteristic of the Adult dataset is its inherent imbalance. Specifically, a substantial proportion of individuals in the dataset have incomes below \$50k (around 75.4\%). The dataset contains several sensitive attributes such as race and gender. We opt to use gender as the sensitive feature in this evaluation. This is also imbalanced with roughly 67.3\% of the data being male. Such imbalances could mislead naive classifiers into an unwanted bias towards the dominant class. Both the gender feature and target income variable are binary. We attempt to eliminate disparities in income predictions across gender groups.

\subsection{COMPAS Dataset}
The Correctional Offender Management Profiling for Alternative Sanctions (COMPAS) dataset became notably popular following an investigation by ProPublica in 2016 \cite{dieterich2016}. COMPAS is a risk assessment tool used in the U.S. legal system to assess the likelihood that a defendant will re-offend. Each instance in the dataset contains 12 features like age, gender, criminal history, and risk scores. The primary task is to predict if an individual will re-offend within two years.

ProPublica's analysis notably highlighted racial disparities in the predictions, where African-American defendants were more likely to be falsely classified as high risk compared to white defendants. We therefore opt to use race as the sensitive feature. The COMPAS dataset is balanced in terms of both the sensitive feature and target variable. The COMPAS dataset one-hot encodes ethnicity into five categories: African American, Asian, Hispanic, Native American, and Other. Studies often reduce this multi-class feature into a binary one distinguishing only between African-American and all other ethnicities, overshadowing the multi-class complexity. To enable comparison against models that do not have multi-class sensitive feature debiasing capability, we perform experiments with this binary simplification. However, we also evaluate our approach with the true multi-class problem. It is important to note that the COMPAS dataset has been heavily cirticised for its use in fairness research due to its inherent measurement biases and errors, its disconnection from real-world criminal justice outcomes and its lack of consideration for the complex normative issues related to fairness, justice and equality \cite{bao2021s}. We use it here only to support comparison against previous works.

\subsection{Data Preprocessing}
For both datasets, we performed standard preprocessing, mapping categorical features to numerical indices, normalization of continuous variables, and handling missing values by replacing them with -1. We split the datasets into training and test sets in a 175:25 ratio. We also drop redundant features, that are either repeated or with mostly missing values. %We end up with each instance having 12 input features (excluding the sensitive feature and target variable) for both datasets being passed as input to the encoder head.

\subsection{Hyperparameter and Model Training}
We employed a learning rate of $0.005$ and $0.01$ and a batch size of $64$ and $32$ for the Adult and COMPAS datasets respectively. The models were trained for $20$ epochs.

For our hyperparameter sensitivity study, we chose values $\Omega \in \{0...100\}$ and $\beta \in \{0.1...100\}$
%$\omega \in \{0, 0.01, 0.1, 1, 3, 5, 10, 15, 20, 50, 100\}$ and $\beta \in \{0.1, 1, 3, 5, 10, 15, 20, 50, 100\}$
, resulting in $100$ tested combinations of $\Omega$ and $\beta$. All models were trained through $5$ distinct runs, and we report the median as well as standard deviation of their performance across the runs. Given the stochastic nature of neural network training, this ensured robustness in our findings. 
%The median grants a central tendency less prone to outliers compared to a simple mean.

\begin{figure*}[t]
\centering
\includegraphics[width=0.43\textwidth]{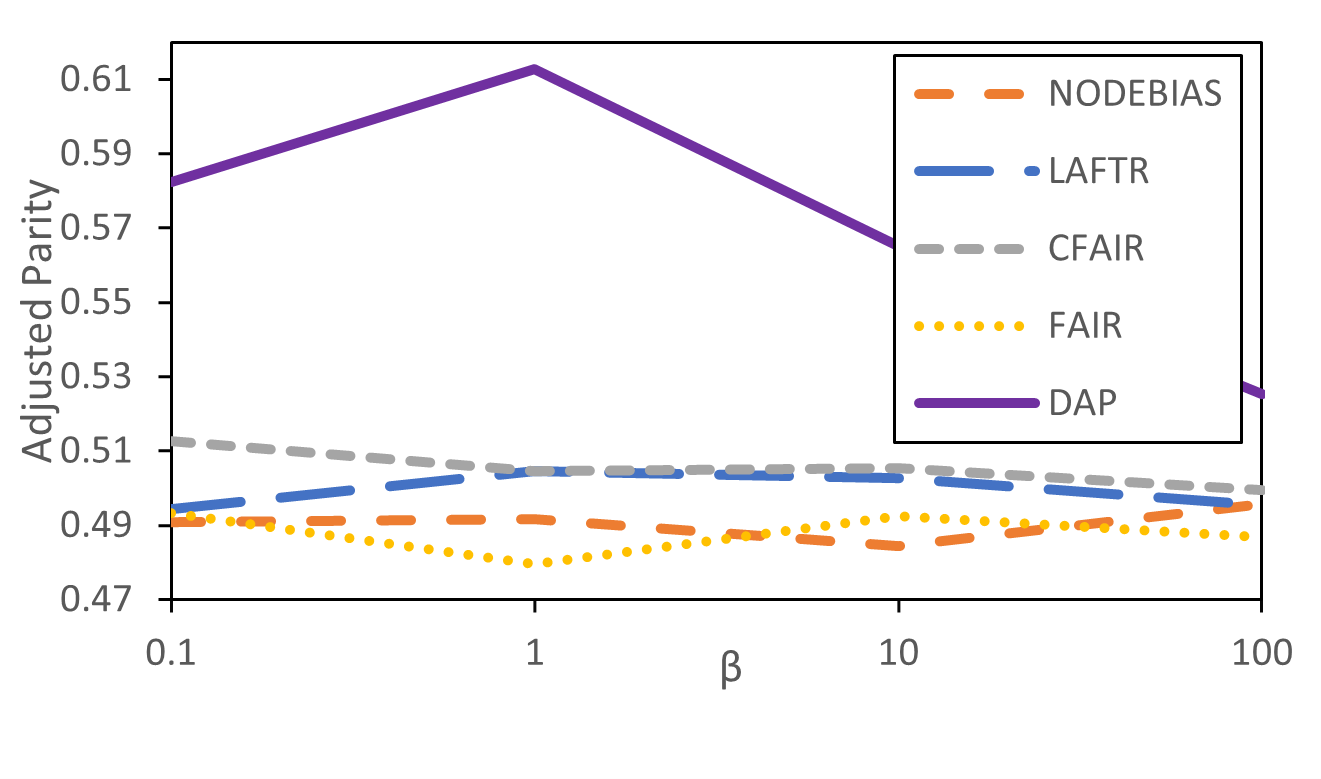}
\includegraphics[width=0.43\textwidth]{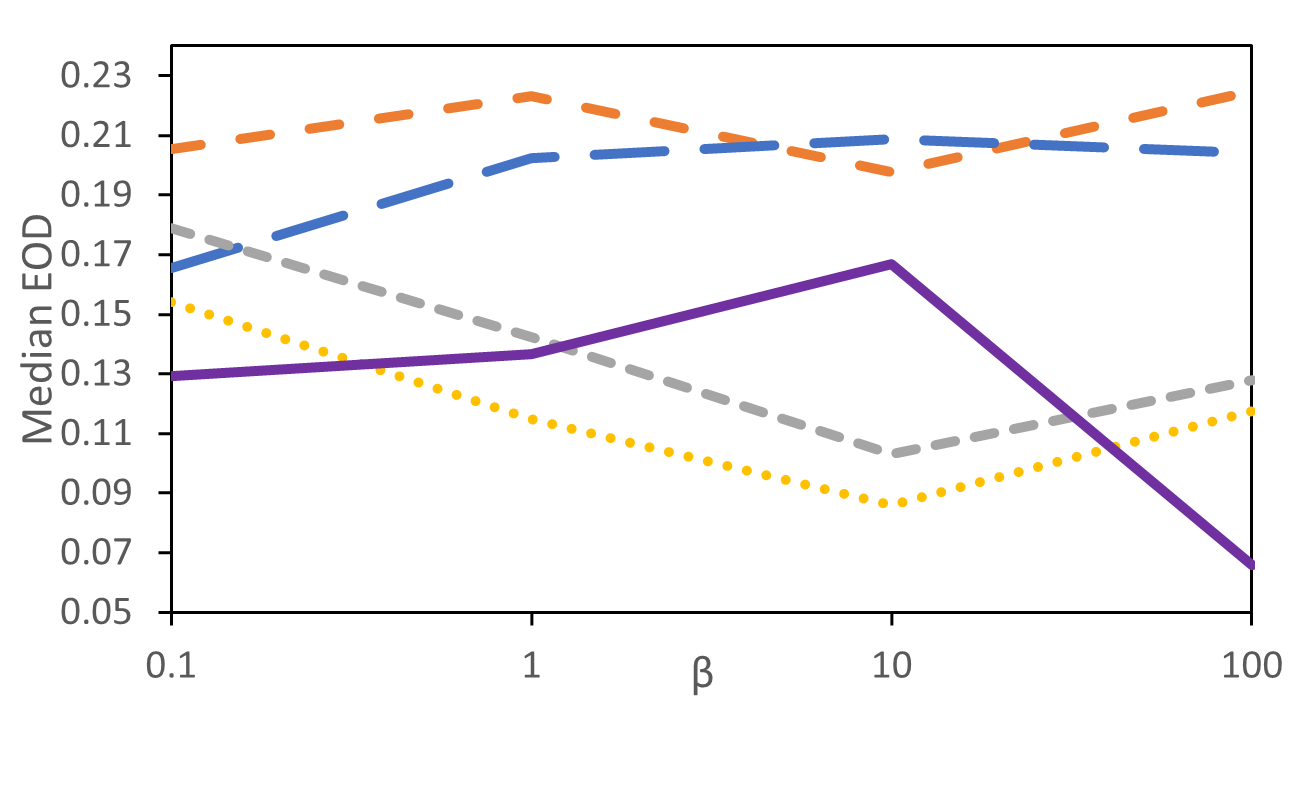} \\
\includegraphics[width=0.43\textwidth]{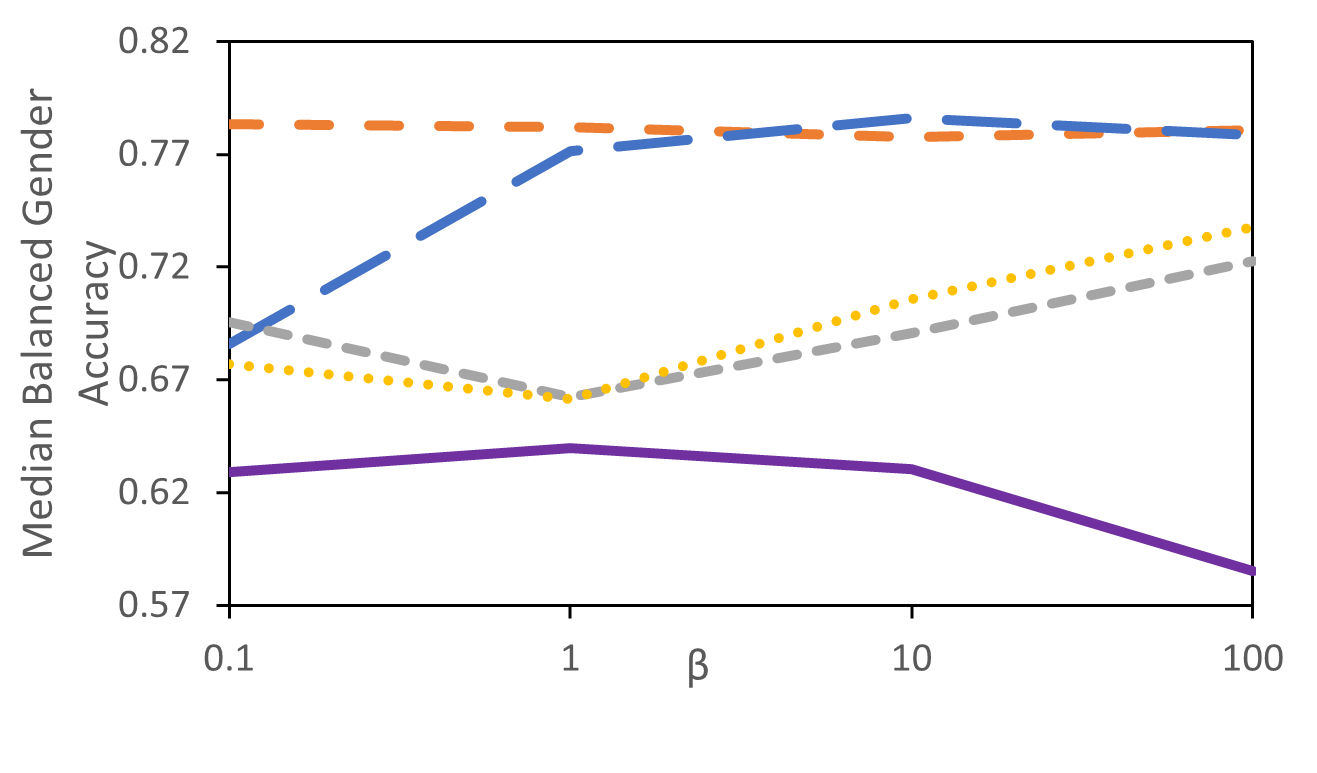}
\includegraphics[width=0.43\textwidth]{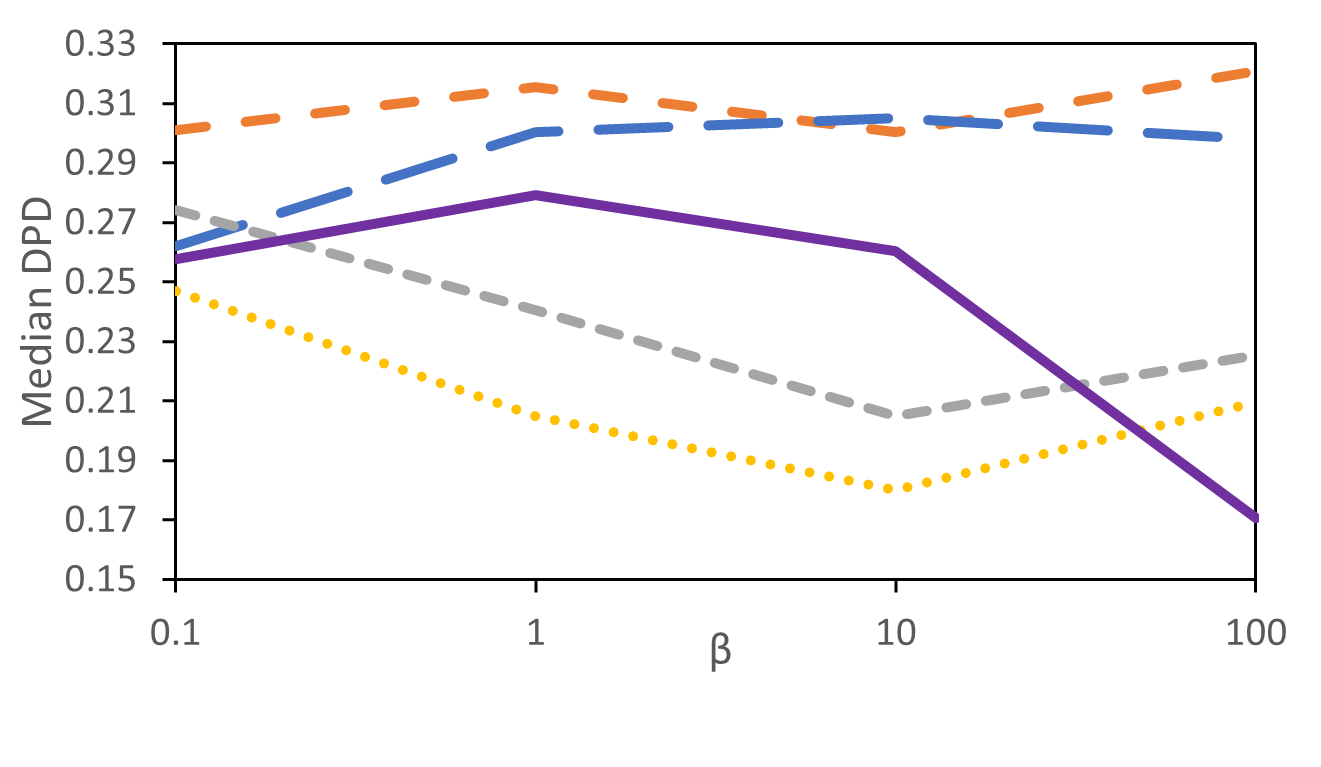}
\caption {Comparing performance and fairness of all 4 models against DAP on Adult dataset. Top-left, top-right, bottom-left and bottom-right graph show how adjusted parity, equalised odds difference (EOD), gender classification accuracy and demographic parity difference (DPD) change with $\beta$. Higher adjusted parity and lower EOD, DPD and gender accuracy are favourable. DAP has higher adjusted parity and lower gender classification accuracy at all $\beta$. Lowest EOD and DPD are obtained by DAP at $\beta$=100.}
\label{fig1}
\end{figure*}

\begin{figure*}[t!]
\centering
\includegraphics[width=0.45\textwidth]{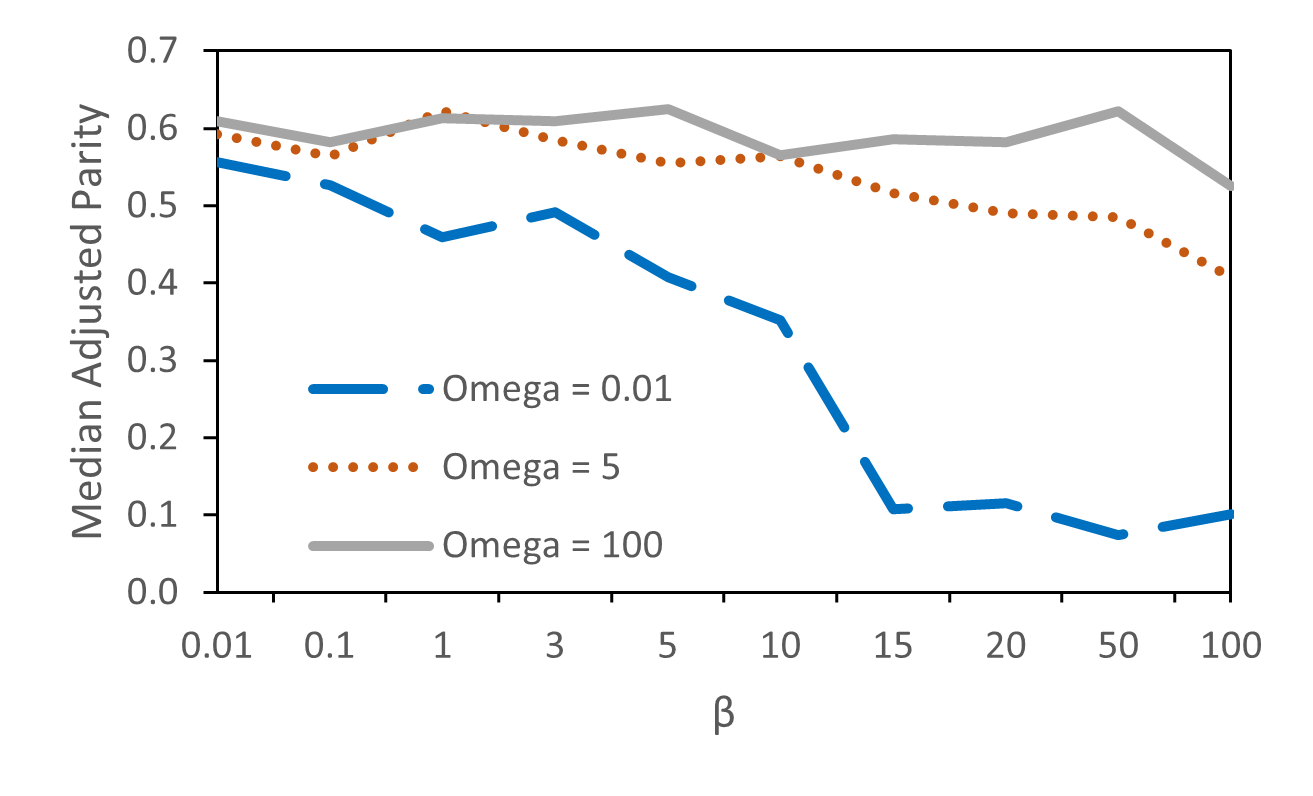}
\includegraphics[width=0.45\textwidth]{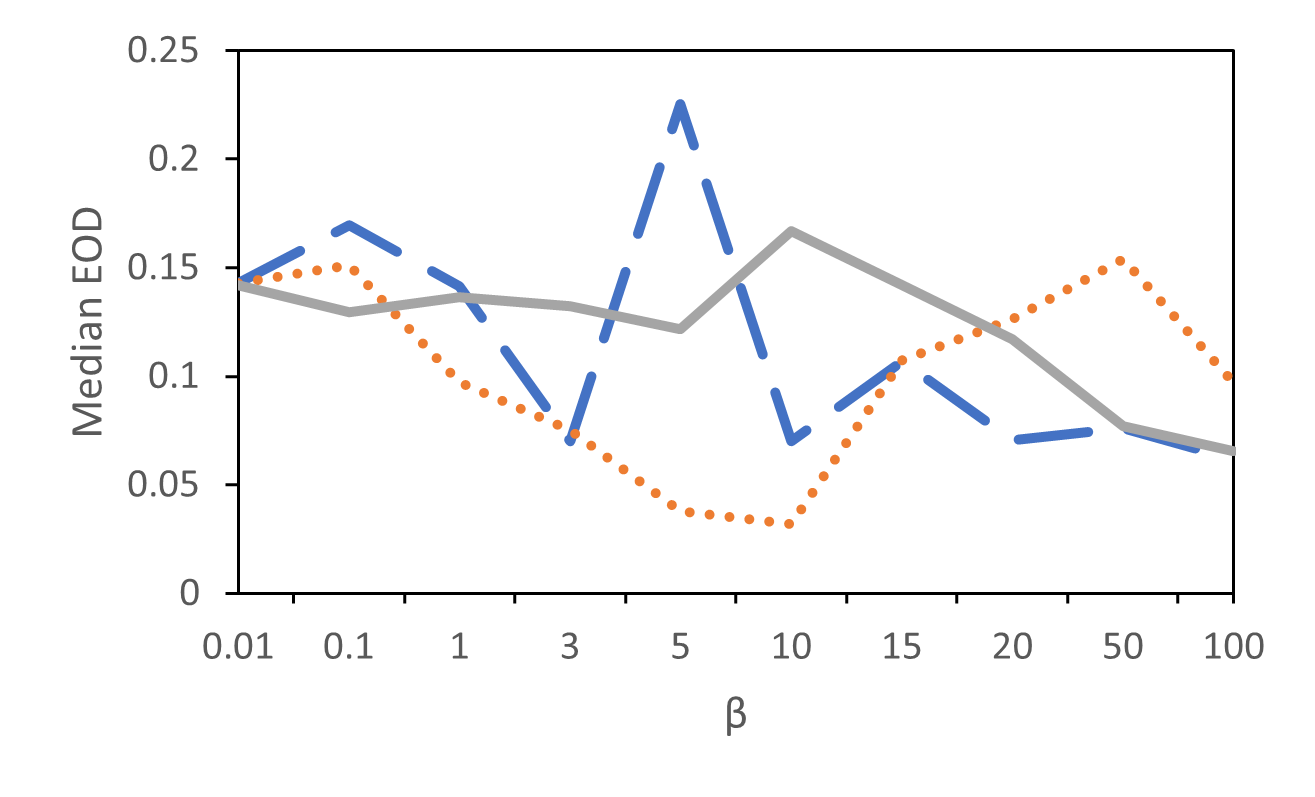} \\
\includegraphics[width=0.45\textwidth]{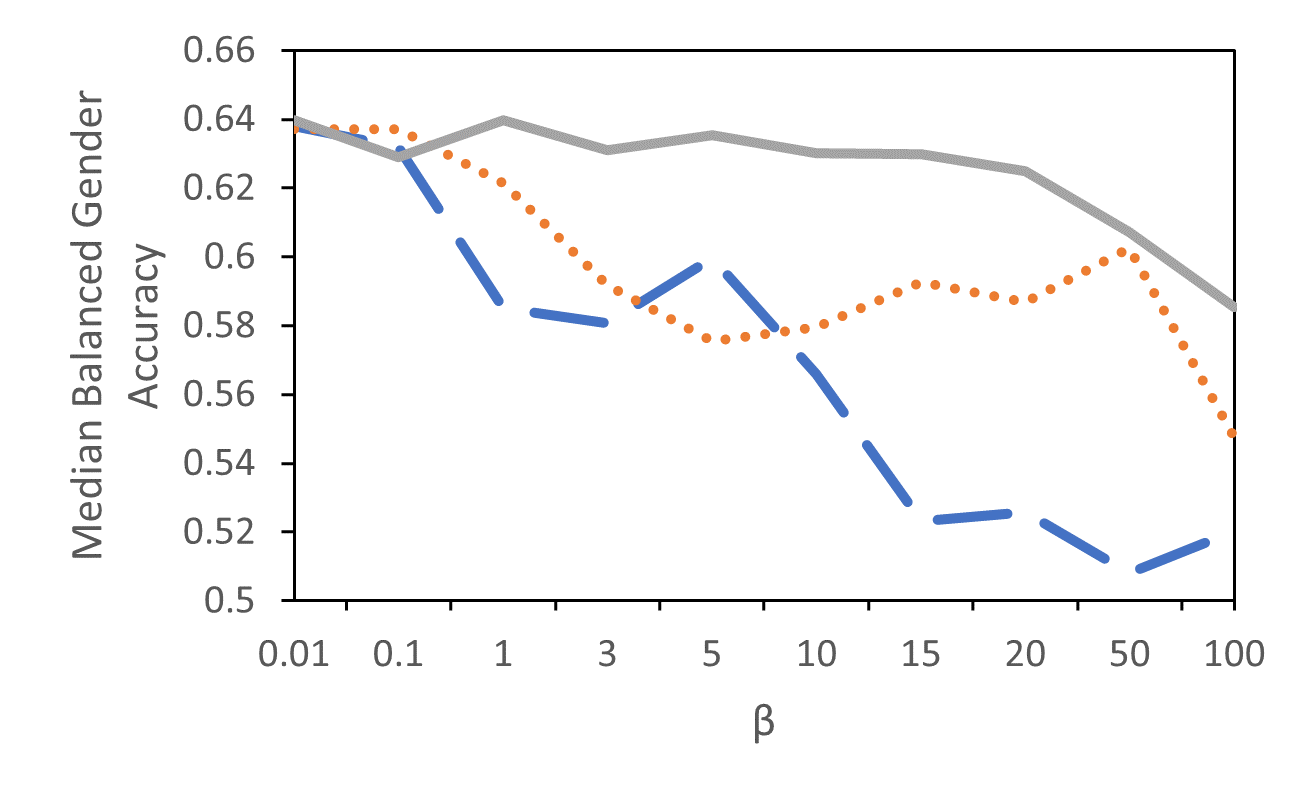}
\includegraphics[width=0.45\textwidth]{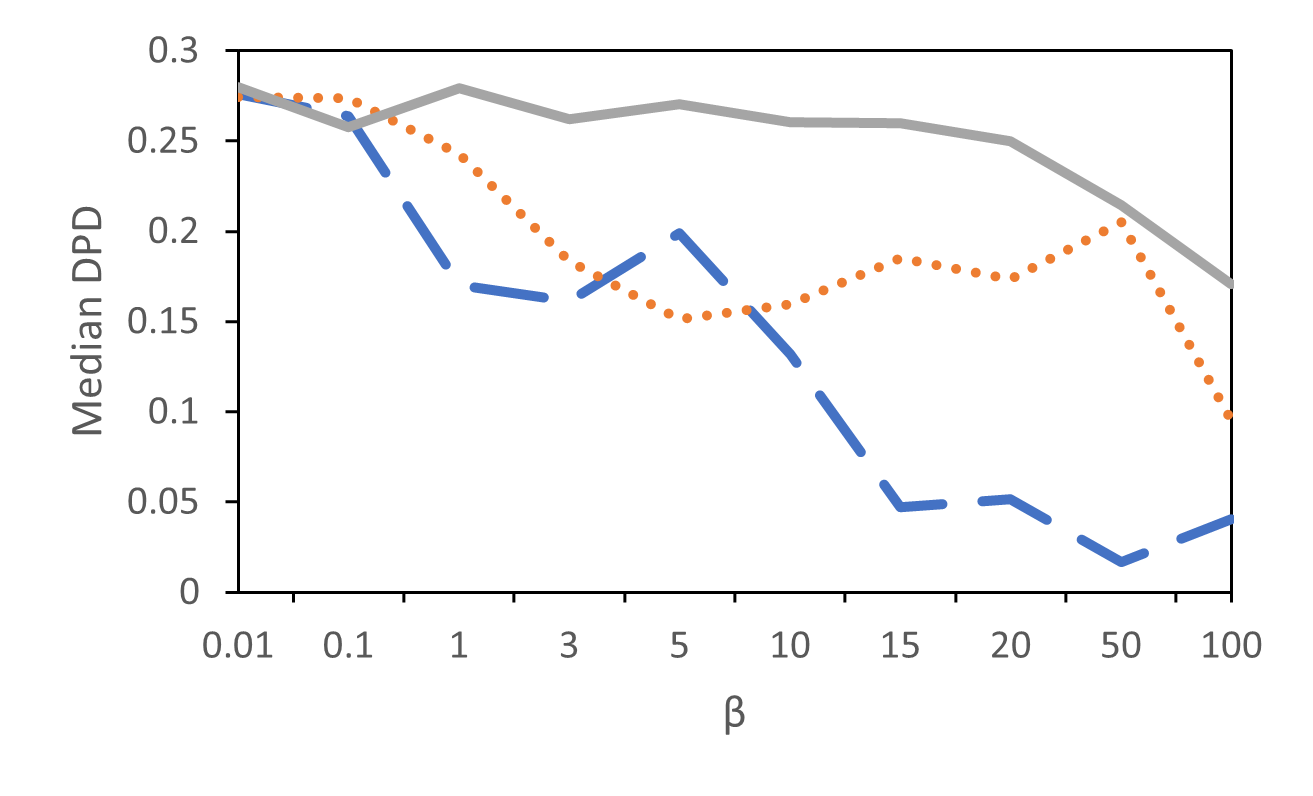}
\caption{Effect of altering $\Omega$ and $\beta$ on adjusted parity (top-left), EOD (top-right), gender accuracy (bottom-left), and DPD (bottom-right) on Adult dataset. Higher adjusted parity and lower EOD, DPD and gender accuracy are favourable. Increasing $\beta$ lowers adjusted parity but improves all other metrics. Effect more pronounced at lower $\Omega$}
\label{fig2}
\end{figure*}

\begin{table*}[t!]
\centering
\begin{tabular}{lcccccl} \hline 
DAP Model  & $\Omega$   & $\beta$ & \begin{tabular}[c]{@{}l@{}} Adjusted \\ Parity\end{tabular} & EOD         & DPD         & \begin{tabular}[c]{@{}l@{}}Gender \\ Accuracy (\%)\end{tabular} \\ \hline 
Balanced   & 100 & 5 & 0.632±0.014                                                & 0.126±0.075 & 0.265±0.028 & 60.9±7.2                                                        \\
Balanced   & 10  & 3 & 0.617±0.005                                                & 0.090±0.005 & 0.233±0.011 & 62.3±4.2                                                        \\
Balanced   & 1   & 1 & 0.593±0.008                                                & 0.052±0.052 & 0.185±0.016 & 59.7±5.5                                                        \\
Balanced   & 15  & 5 & 0.621±0.009                                                & 0.111±0.010 & 0.242±0.018 & 61.1±1.9                                                        \\
Balanced   & 3   & 1 & 0.622±0.006                                                & 0.113±0.038 & 0.244±0.011 & 59.3±4.4                                                        \\
Unbalanced & 100 & 5 & 0.635±0.013                                                & 0.122±0.059 & 0.271±0.025 & 62.5±5.9                                                        \\
Unbalanced & 10  & 3 & 0.618±0.011                                                & 0.096±0.020 & 0.236±0.022 & 61.4±3.3                                                        \\
Unbalanced & 1   & 1 & 0.596±0.010                                                & 0.053±0.033 & 0.193±0.019 & 61.0±4.1                                                        \\
Unbalanced & 15  & 5 & 0.615±0.010                                                & 0.094±0.009 & 0.230±0.020 & 59.9±3.8                                                        \\
Unbalanced & 3   & 1 & 0.619±0.005                                                & 0.114±0.045 & 0.239±0.011 & 60.7±5.2  \\   \hline  \hline         
\end{tabular}
\caption{Comparing the adjusted parity, EOD, DPD, gender accuracy when balanced and unbalanced accuracies are used during training. This table shows a few high performing combinations of $\beta$ and $\Omega$ on the Adult dataset. No significant difference is observed between Balanced and Unbalanced.
}
\label{table1}
\end{table*}

\begin{figure*}[t]
\centering
\includegraphics[width=0.43\textwidth]{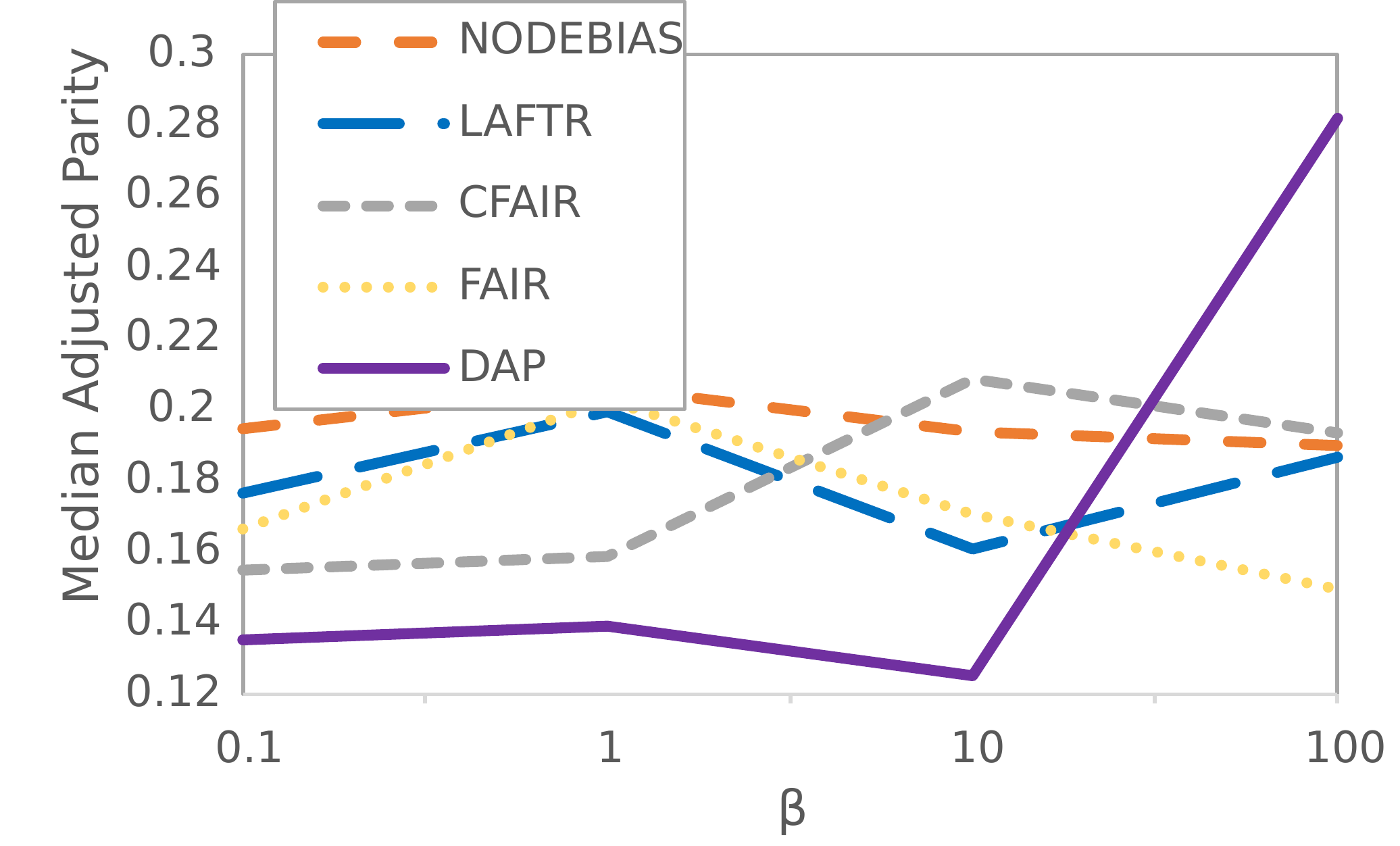}
\includegraphics[width=0.43\textwidth]{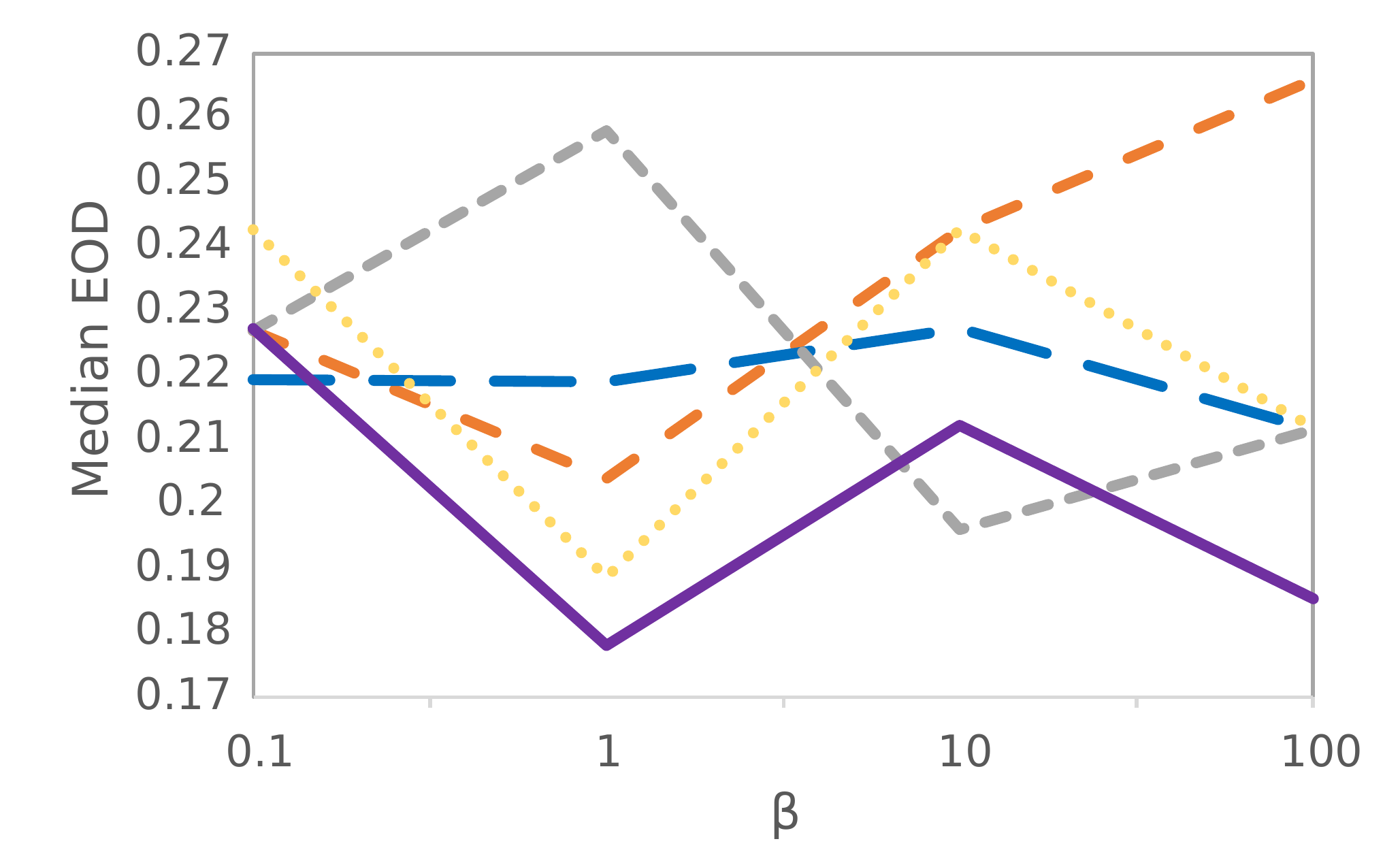} \\
\includegraphics[width=0.43\textwidth]{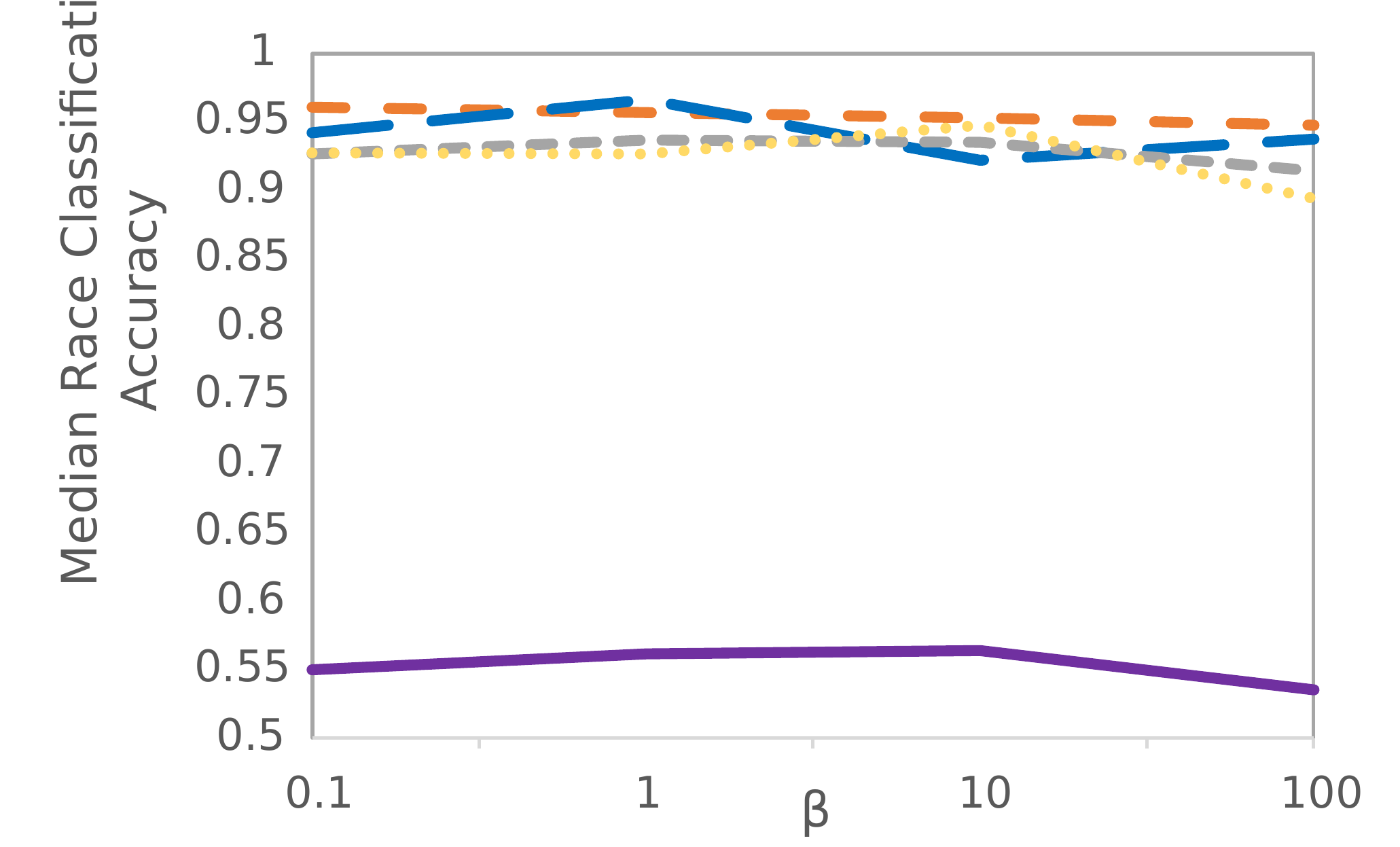}
\includegraphics[width=0.43\textwidth]{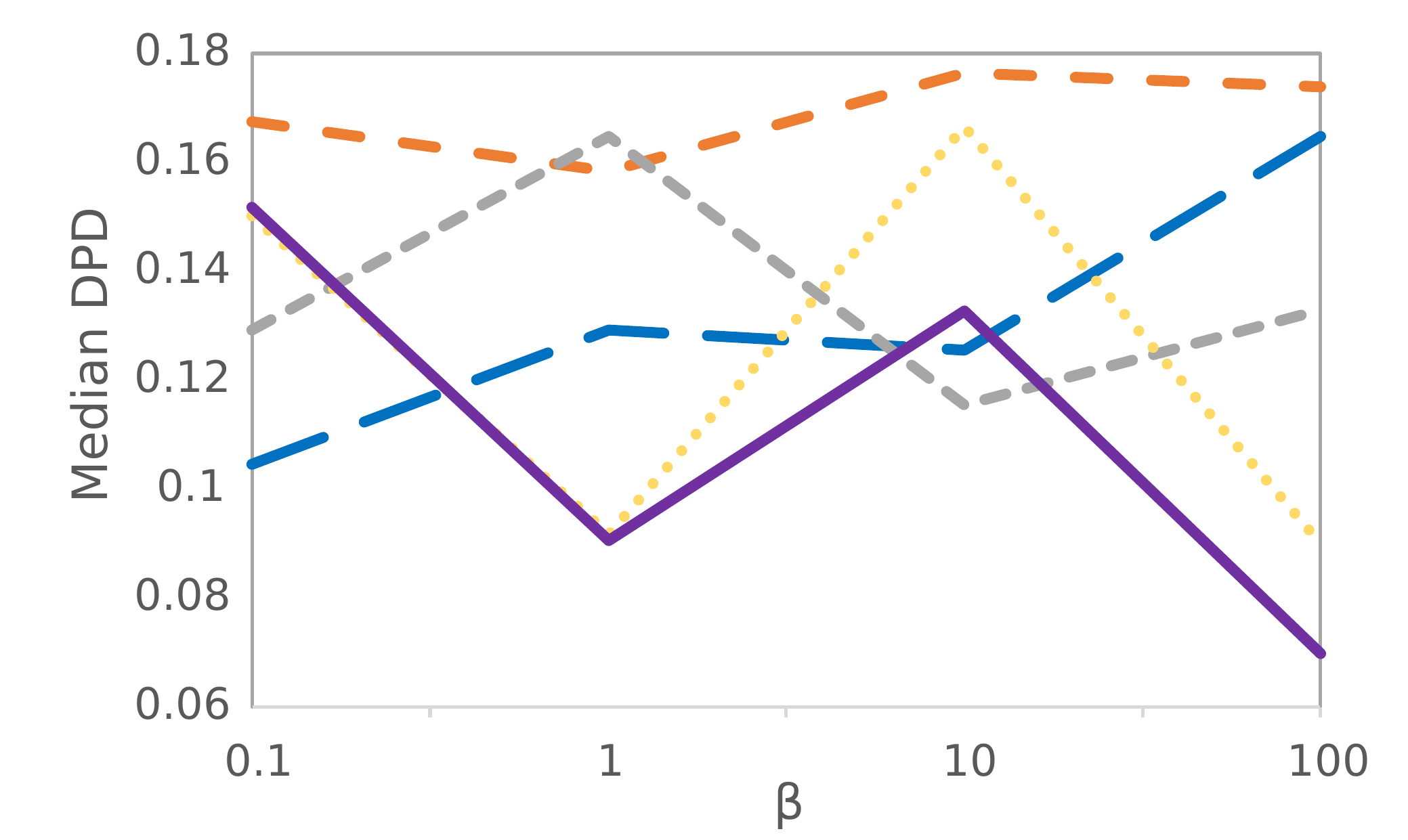}
\caption{Comparing all 4 models performance and fairness against DAP on the COMPAS dataset. Top-left, top-right, bottom-left and bottom-right graph show how adjusted parity, EOD, race classification accuracy and DPD change with $\beta$. Higher adjusted parity and lower EOD, DPD and gender accuracy are favourable. DAP has lower gender classification accuracy at all $\beta$. Highest adjusted parity and lowest EOD and DPD are obtained by DAP}
\label{fig3}
\end{figure*}
\subsection{Evaluation Framework}
To evaluate our models' performance, we trained 2 balanced random forest classifiers to predict the sensitive and target features from the encodings in the testing phase. This allowed us to measure balanced classification accuracies on the task variable and sensitive feature from the embeddings produced during testing. On the fairness front, we obtain fairness metrics like demographic parity and equalised odds, using the predicted target from the random forest classifier. We also obtain an adjusted parity metric for comparing models and selecting best performing hyperparameters.

We compare our model with CFAIR, LAFTR and FAIR (ALFR), which we covered in \ref{sec:background}. We also compare against  NODEBIAS which is reference network devoid of fairness constraints. The implementation for these models was adapted from \cite{taejunkim}. We evaluate them using the same evaluation protocol as for our model with balanced random forest classifier to obtain target and sensitive feature classification accuracy, DP, EO and adjusted parity metrics. We also contrast our DAP model using balanced soft accuracies against a variant using unbalanced soft accuracies.

\begin{figure*}[t]
\centering
\begin{minipage}{0.43\textwidth}
    \centering
    \textbf{Adult} \par\medskip
    \includegraphics[width=\textwidth]{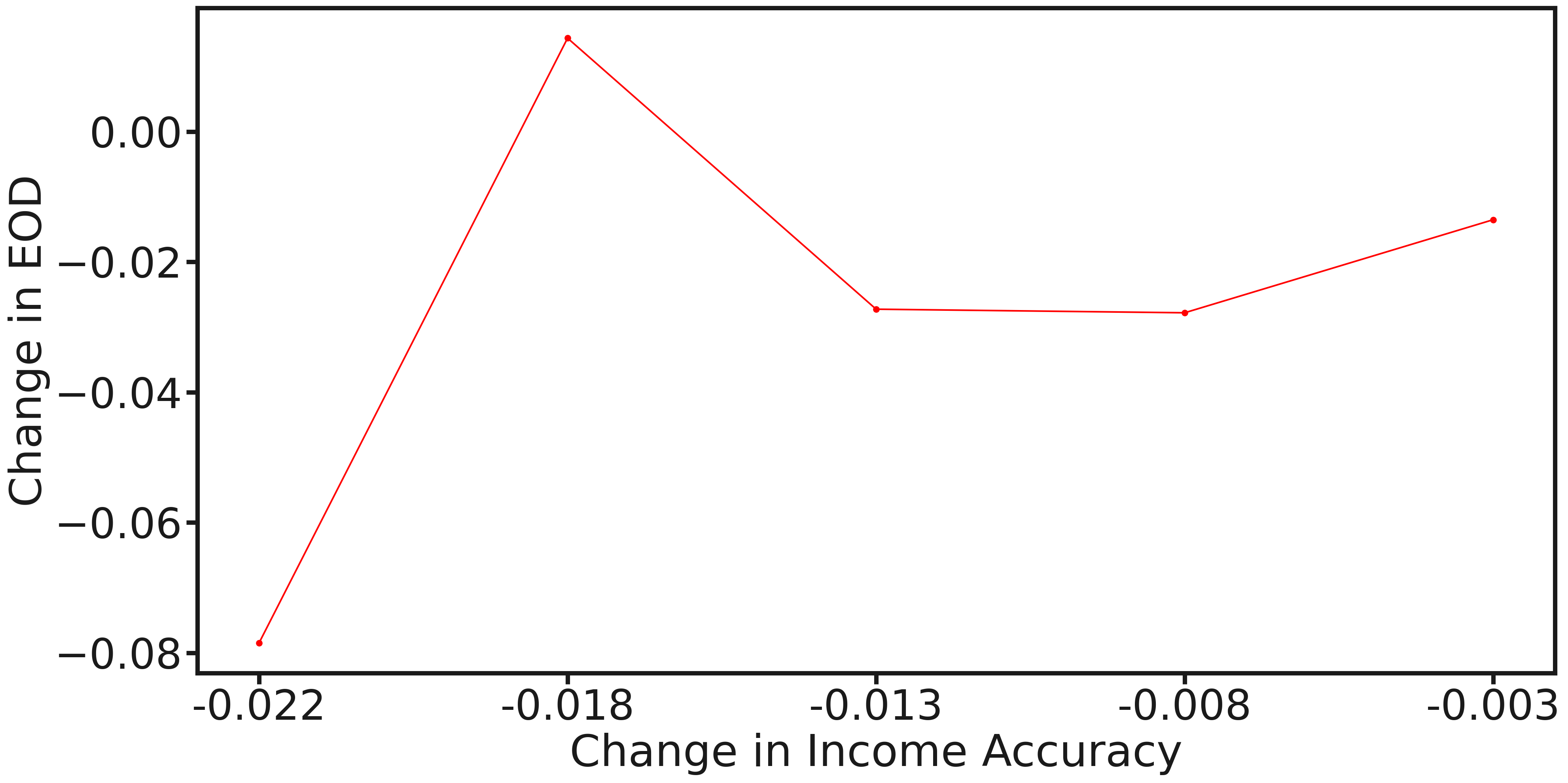} \\
    \includegraphics[width=\textwidth]{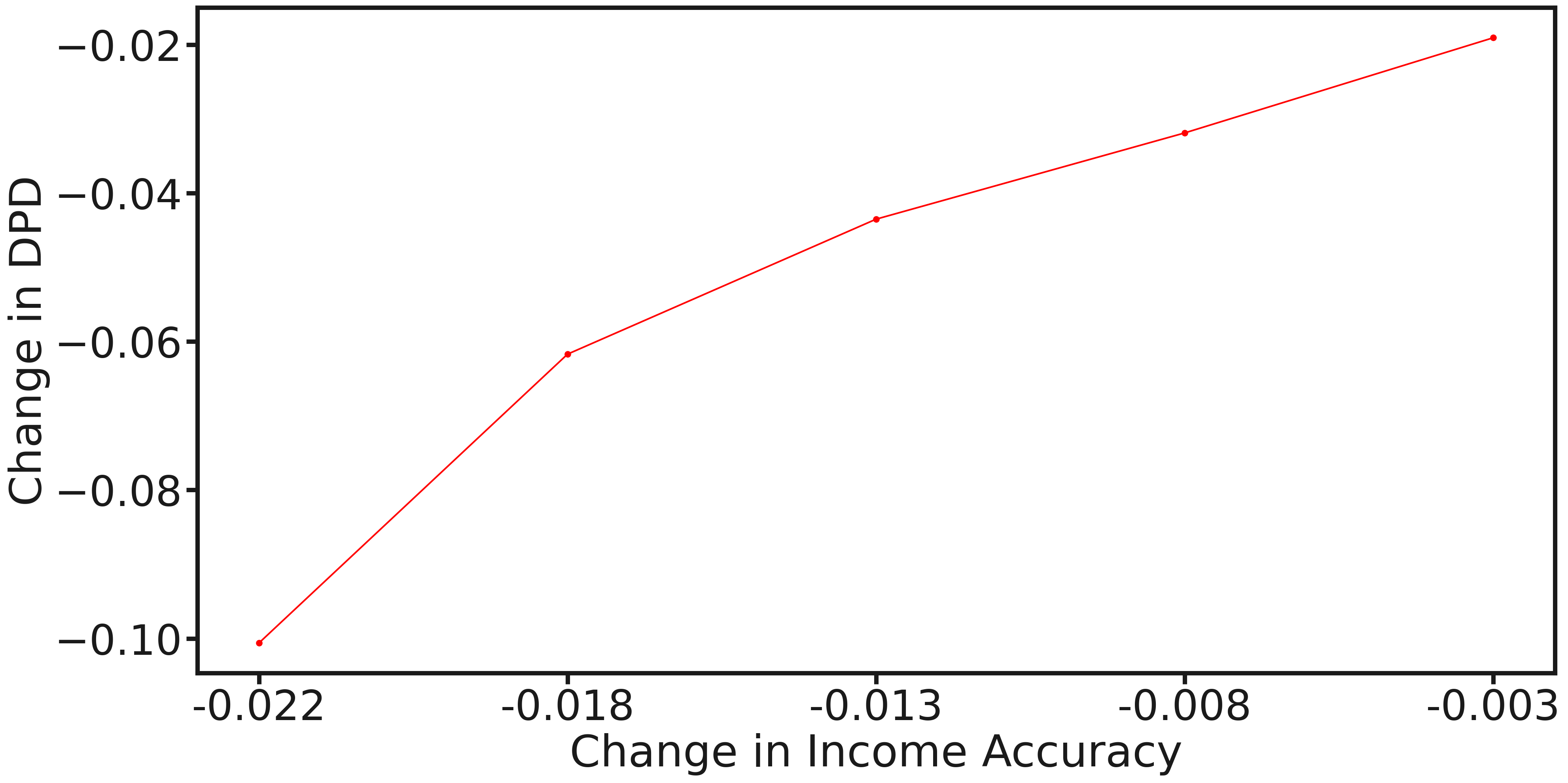} \\
    \includegraphics[width=\textwidth]{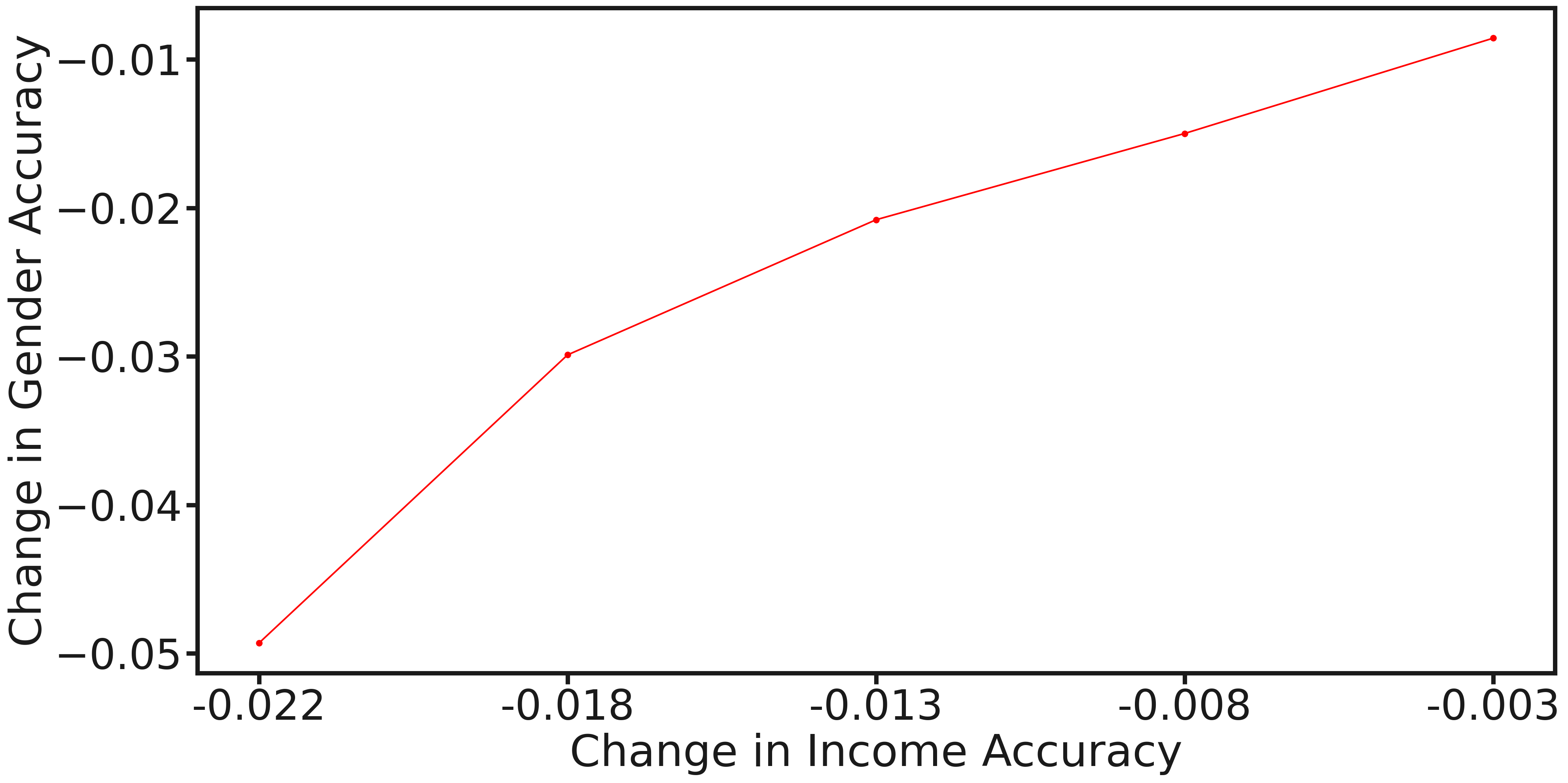}
\end{minipage}
\hspace{0.05\textwidth} % Add horizontal space between columns
\begin{minipage}{0.43\textwidth}
    \centering
    \textbf{COMPAS} \par\medskip
    \includegraphics[width=\textwidth]{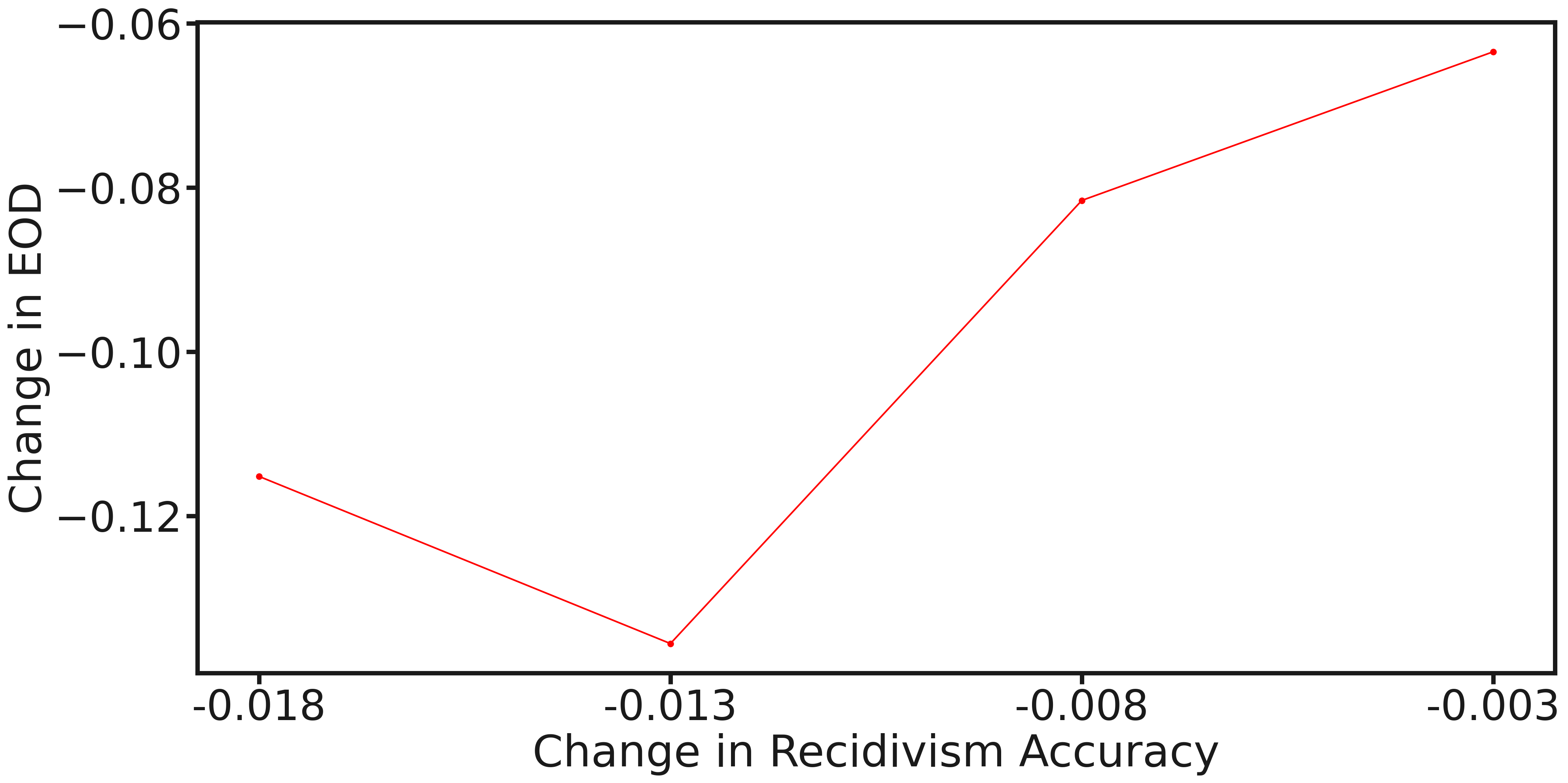} \\
    \includegraphics[width=\textwidth]{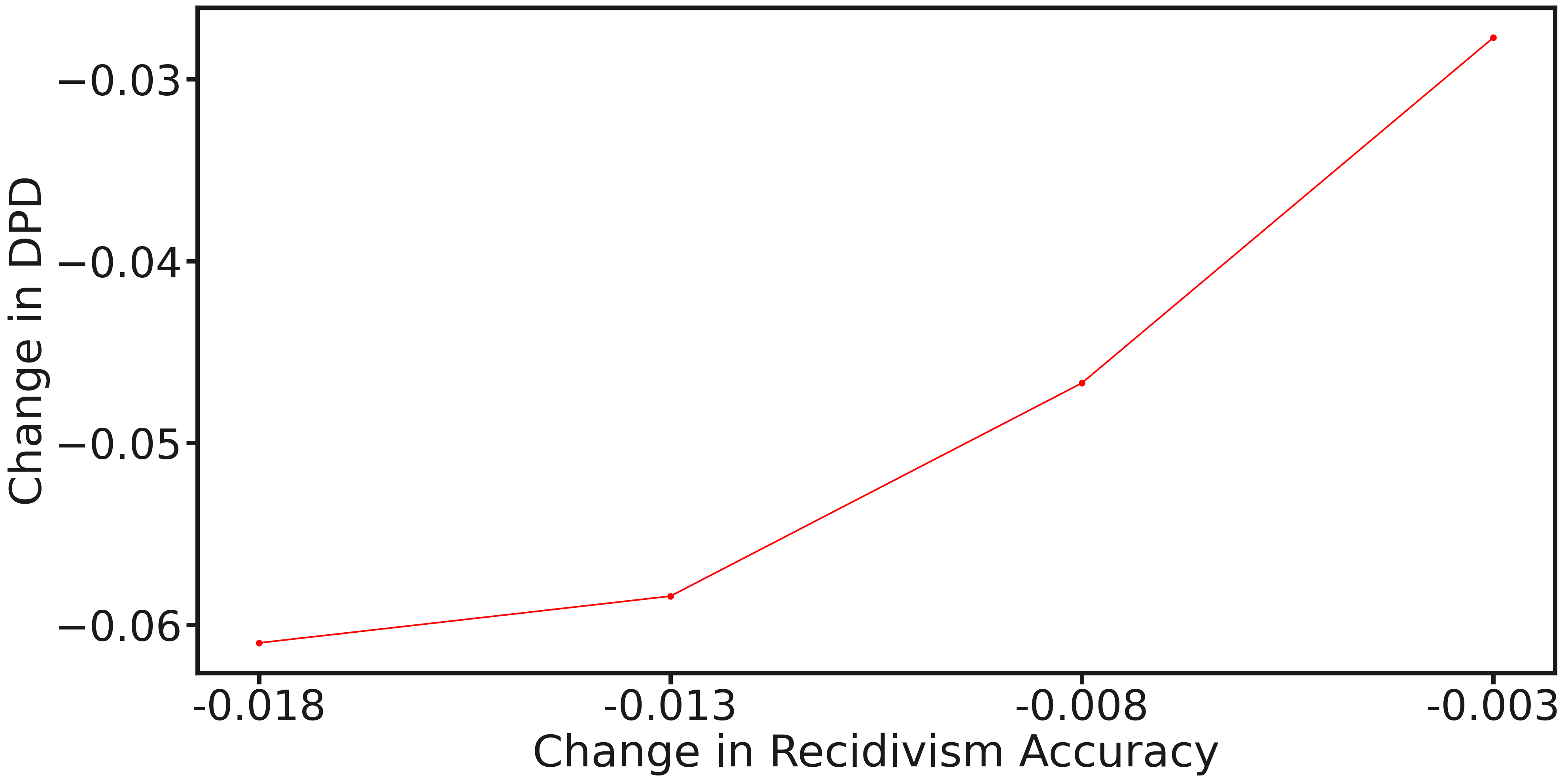} \\
    \includegraphics[width=\textwidth]{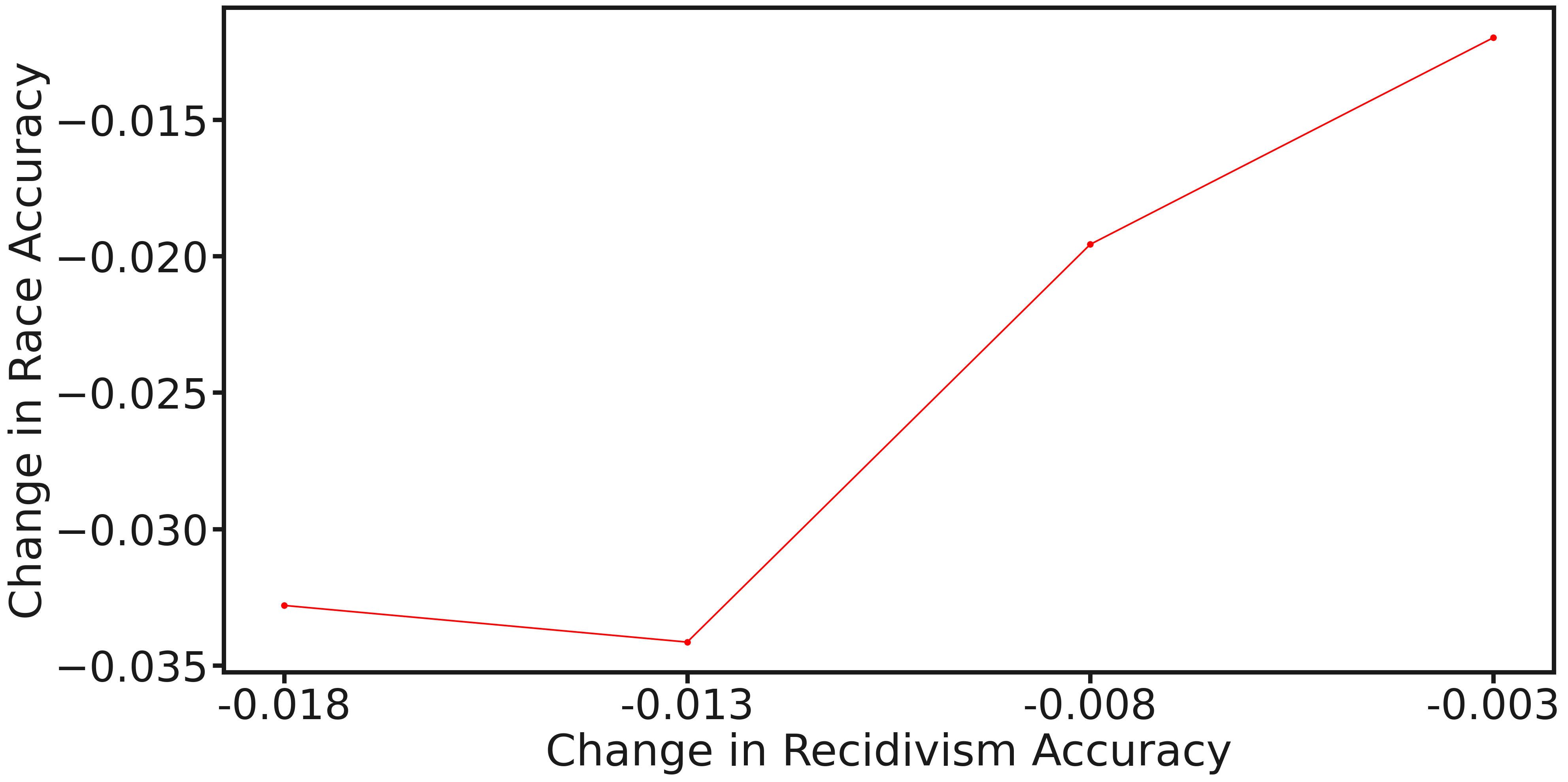}
\end{minipage}
\caption{Showing how the fairness metrics vary with changes in task accuracy from baseline values (Income Accuracy: 0.8294 for Adult, Recidivism Accuracy: 0.584 for COMPAS; EOD: 0.1429/0.310; DPD: 0.2759/0.173; Gender/Race Accuracy: 0.637/0.581) for each metric for Adult (left) and COMPAS (right). Changes were binned into 0.005 intervals, and averages for EOD, DPD, and gender/race accuracy were computed for each interval. Obtaining the largest negative change in EOD, DPD and sensitive feature accuracy for the least drop in task accuracy is favorable.}
\label{fig5}
\end{figure*}

\label{sec:results}
As depicted in Figure \ref{fig1}, DAP consistently outperforms NODEBIAS, LAFTR, CFAIR, and FAIR in terms of adjusted parity and gender classification balanced accuracy on the Adult dataset. Specifically, DAP demonstrates a superior adjusted parity (which is notably higher) and a more favorable gender classification balanced accuracy (which is significantly lower). When benchmarked against EO and DP metrics, DAP either achieves equivalent or better than the other 4 models. The lowest EOD and DPD are achieved with DAP. With \( \beta = 100 \), DAP obtains 44.1\% and 18.6\% lower EOD and DPD than the next best performer FAIR.

Figure \ref{fig2} delves into the effects of tweaking $\beta$ and $\Omega$ on the performance of DAP. When $\Omega$ is held constant and $\beta$ is gradually increased, there is a notable decrease in the adjusted parity. Concurrently, other fairness metrics such as EOD, DPD, and gender classification balanced accuracy witness marked improvements. This effect is particularly evident at lower $\Omega$ values. Conversely, when $\beta$ remains static and $\Omega$ is increased, the outcomes typically include a surge in adjusted parity, gender classification balanced accuracy, and DPD metrics. Interestingly, EOD doesn't seem to follow a discernible trend in relation to changing $\Omega$ values. It's important to highlight that specific pairings of $\beta$ and $\Omega$ can optimize EOD values, indicating the delicate interplay between these parameters.

In an evaluative comparison between using balanced and unbalanced soft accuracies during training, Table \ref{table1} underscores that there are negligible differences in the adjusted parity. Moreover, the differences in adjusted parity, gender classification accuracy, EOD, and DPD are all insignificant as emphasized by the overlap of the standard deviations.

Figure \ref{fig3} presents the experimental outcomes on the COMPAS dataset. Here DAP exhibits consistent improvements over previous state-of-the-art. In particular for high values of  \( \beta \) (meaning a high fairness weighting), DAP outperforms all competing approaches. 
%It increases adjusted parity and reduces DPD, EOD, and race classification accuracy (all with reduced values). More specifically, 
DAP achieves an adjusted parity that is improved by 45.9\% and an EOD that is reduced by 12.4\% when compared with its nearest competitor, CFAIR. Similarly, it registers a 22.5\% improvement in DPD and a substantial 40.1\% reduction in race classification accuracy when compared to FAIR, the latter being the second-best performer for these metrics. Unlike with the Adult dataset, the performance of DAP on COMPAS does not seem to be very sensitive to the value of $\Omega$. Because the performance is roughly similar across all values, the results are omitted here and can be found in section Appendix A. However, they can be found in the supplementary material.

Figure \ref{fig5} demonstrates the interplay between improving fairness metrics and declining task performance. Ideally, we aim to minimize the impact on task accuracy while maximizing the reduction in fairness-related metrics. With less than a 2.5\% decrease in income classification accuracy, a reduction of 0.08, 0.10, and 4.93\% is achieved on EOD, DPD, and gender classification accuracy, respectively, on the Adult dataset. Similarly, a decline of less than 2\% in recidivism accuracy results in a decrease of 0.14, 0.06, and 3.41\% in EOD, DPD, and race classification accuracy on the COMPAS dataset.

\begin{figure}[t]
\centering
\includegraphics[width=0.6\columnwidth]{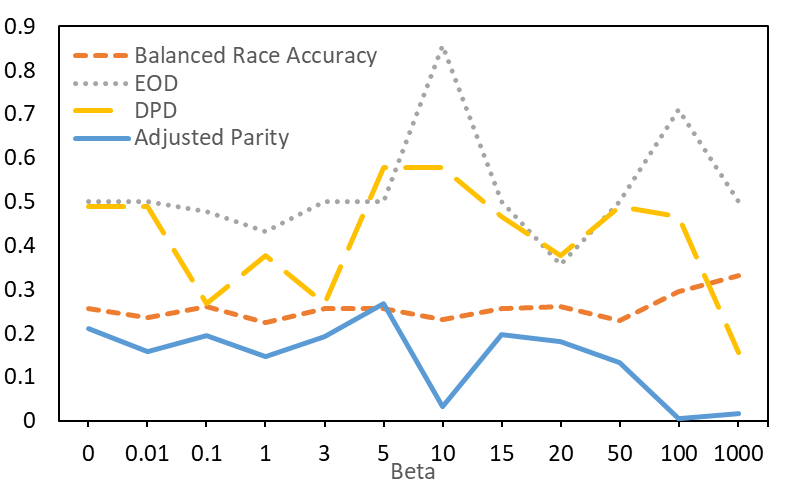}
\caption{Demonstrating the performance of DAP with multi-class sensitive features at $\Omega$=20. DPD and adjusted parity decrease and race classification accuracy approaches 0.33 with increasing $\beta$, as desired. EOD shows no significant trend. %Higher adjusted parity and lower DPD and EOD are favourable. Race classification accuracy close to 0.33 (random chance) is preferred.
}
\label{fig4}
\end{figure}
%Figure \ref{fig4} illustrates the effectiveness of DAP when applied to multi-class sensitive attributes. With increasing $\beta$, EOD remains constant, starting and ending at 0.5, with no percentage change and an insignificant negative trend (slope: -0.0000132, p-value: 0.9292). In contrast, DPD experiences a significant decrease of 68.18\%, dropping from 0.4889 to 0.1556, alongside a statistically significant negative trend (slope: -0.0002808, p-value: 0.0350). Balanced race classification accuracy shows an increase of 28.82\%, rising from 0.2570 to 0.3311, with a statistically significant positive trend (slope: 0.0000839, p-value: 0.0018). With three race classes, random chance balanced accuracy would be 0.33. An accuracy closer to this value indicates less race information present in the embeddings. Finally, increases slightly upto 0.27 at $\beta$=5, but overall shows a substantial decrease of 92.14\%, falling from 0.2100 to 0.0165.
Figure \ref{fig4} illustrates the effectiveness of DAP for multi-class sensitive attributes. As $\beta$ increases, EOD remains constant, starting and ending at 0.5 with no significant negative trend. In contrast, DPD significantly decreases by 68.18\%, dropping from 0.4889 to 0.1556. Balanced race classification accuracy improves by 28.82\%, rising from 0.2570 to 0.3311, nearing the random chance level of 0.33, indicating reduced race information in the embeddings. Finally, adjusted parity increased slightly up to 0.27 at $\beta=5$, but overall shows a substantial drop of 92.14\%, from 0.2100 to 0.0165.

%DAP significantly improves over previous SOTA in terms of race classification accuracy. Here, it consistently surpasses the other four models by maintaining a race classification accuracy that's lower by approximately 40\%, highlighting the model's robustness in this particular aspect. The optimal performance metrics of DAP are attained through specific combinations of \( \beta \) and \( \Omega \).

\section{Conclusion}
\label{sec:conclusion}
The results unequivocally position DAP as a highly effective approach for achieving fairness across different datasets, significantly outclassing established models like FAIR, CFAIR, and LAFTR. 
These adversarial approaches can prove challenging to train but DAP remains stable, even for complex under-explored problems like those with multi-class sensitive features.
%This distinction becomes particularly important when considering the mechanism of these models. While FAIR, CFAIR, and LAFTR primarily leverage adversarial components that can be unstable and challenging to train, DAP presents an alternative.

%We also demonstrated DAP's effectiveness when working with multi-class sensitive features on the COMPAS dataset. DAP's performance is on par with, or even surpasses, its effectiveness on binary sensitive features. This capability is significant as it inherently integrates multi-class sensitive features without necessitating modifications to the model architecture.

We formulate a differentiable variant of the adjusted parity metric, which includes only adaptively weighted positive learning signals with no adversarial tension. At its core, this involves the innovative use of the ``Soft Balanced Accuracy'' to provide a metric which is smoothly differentiable and agnostic to dataset biases (either in terms of sensitive characteristics or end-task labels).
Unlike other non-adversarial approaches to fairness, DAP also does not suffer from degenerate solutions. The metric cannot be satisfied by performing equally poorly across all sensitive domains.

In terms of the limitations of DAP, there is some sensitivity to the hyperparameters $\beta$ and $\Omega$ as illustrated in our sensitivity studies. The effects of $\beta$ and $\Omega$ are nuanced and interconnected, making it necessary to calibrate them specifically depending on the application and dataset characteristics.
%While increasing $\beta$ notably diminishes the adjusted parity, it simultaneously enhances other fairness metrics. However, the effects are more pronounced at lower $\Omega$ values. The subtle interplay between these parameters makes it necessary to calibrate them specifically depending on the application requirements and dataset characteristics.

Future work in this area should focus on techniques to automatically calibrate the hyperparameters for a given problem. There may be some benefits in developing a scheme to adapt the hyperparameters throughout the training process based on current performance. This would be a natural extension of DAP's current implicit approach to adaptive loss weighting based on fairness measures.

% In the unusual situation where you want a paper to appear in the
% references without citing it in the main text, use \nocite
\nocite{langley00}

\bibliography{neurips_2024}
\bibliographystyle{plainnat}

%%%%%%%%%%%%%%%%%%%%%%%%%%%%%%%%%%%%%%%%%%%%%%%%%%%%%%%%%%%%%%%%%%%%%%%%%%%%%%%
%%%%%%%%%%%%%%%%%%%%%%%%%%%%%%%%%%%%%%%%%%%%%%%%%%%%%%%%%%%%%%%%%%%%%%%%%%%%%%%
% APPENDIX
%%%%%%%%%%%%%%%%%%%%%%%%%%%%%%%%%%%%%%%%%%%%%%%%%%%%%%%%%%%%%%%%%%%%%%%%%%%%%%%
%%%%%%%%%%%%%%%%%%%%%%%%%%%%%%%%%%%%%%%%%%%%%%%%%%%%%%%%%%%%%%%%%%%%%%%%%%%%%%%
\newpage
\appendix
\section{Derivation of maximum standard deviation.}

We define our Differential Adjusted Parity as
\vspace{-0.1cm}\begin{equation}
    \Delta_{\text{adj}} = (\bar{S} - S^{r})\left(1 - \frac{\sigma}{\gamma}\right)
\label{eq:diffpar}
\end{equation}
where $S^r$ is the baseline accuracy of a random predictor, $\bar{S}$ is the average Soft Balanced Accuracy across sensitive domains, $\sigma$ is the standard deviation of SAB across these domains and $\gamma$ is the maximum possible standard deviation. The term $\gamma$ serves to normalise the metric between [0,1].

For any even number of domains ($N$) $\gamma=0.5$. However for an odd number of domains 
\vspace{-0.1cm}\begin{equation}
    \gamma = \sqrt{\frac{1}{4} \left(1 - \frac{1}{N^2}\right)}.
    \label{eq:gamma2}
\end{equation}
Below we present the derivation of this rule.

The standard deviation of values, \( \sigma \), is defined as:
\vspace{-0.1cm}\begin{equation}
    \sigma = \sqrt{\frac{1}{N}\sum_{s \in S} (s - \bar{S})^2}
\end{equation}
where \( \bar{S} \) is the mean of the values in \( S \), representing the mean of the soft balanced accuracies.

The formulation for the maximum standard deviation, \( \gamma \), is:
\vspace{-0.1cm}\begin{equation}
    \gamma = \arg \max_{S \in \mathbb{R}^N} \sqrt{\frac{1}{N}\sum_{s \in S} (s - \bar{S})^2} 
     \quad  \quad \text{s.t. } s \in [\alpha..\Omega]
\end{equation}
where $\alpha$ and $\Omega$ are the upper and lower bounds on the values of $s$. In other words we use soft balanced accuracy vector $S$ which leads to maximal standard deviation from the mean.

We can see by inspection that when the values of $s$ are bounded, we achieve maximial standard deviation by making \( \bar{S} \) as centered as possible within the range, while all the values in $s$ are on the extremes of the range. When \( N \) is even, this can be perfectly achieved by placing half of the items at each end. In such a scenario, \( \bar{S} = \sigma = \frac{\alpha + \Omega}{2} \). Hence, if the soft balanced accuracies are bounded between [0..1] then the maximum standard deviation for any even number of sensitive domains is 0.5.

However, for odd \( N \), the mean will necessarily be slightly off-center from the range, leading to a lower maximal \( \gamma \). Considering that \( \mathrm{floor}(N/2) \) items are placed at \( \alpha \) and \( \mathrm{ceil}(N/2) \) items are placed at \( \Omega \), the mean can be derived as:
\begin{align}
    \bar{S} &= \frac{\alpha\mathrm{floor}\left(\frac{N}{2}\right) + \Omega\mathrm{ceil}\left(\frac{N}{2}\right)}{N} \\
    & = \frac{\alpha\frac{N-1}{2} + \Omega\frac{N+1}{2}}{N}
\end{align}

Similarly, the summation inside the definition of $\sigma$ can be split into 2 parts and resolved
\begin{align}
    \gamma & = 
    \sqrt{\frac{1}{N}\left(
    \sum_{i=1}^{\frac{N-1}{2}}
    (\bar{S} - \alpha)^2
    +
    \sum_{i=1}^{\frac{N+1}{2}}
    (\bar{S} - \omega)^2
    \right)}
    \\ & = 
    \sqrt{\frac{1}{N}\left(
    \frac{N-1}{2}
    (\bar{S} - \alpha)^2
    +
    \frac{N+1}{2}
    (\bar{S} - \omega)^2
    \right)}
    \label{eq:stdev}
\end{align}

Given that the minimum normalized accuracy, \( \alpha \), is 0 and the maximum normalized accuracy, \( \Omega \), is 1, the simplified mean is:
\begin{align}
    \bar{S} &= \frac{1+\frac{1}{N}}{2} \\
    &= \frac{1}{2} + \frac{1}{2N}
\label{eq:mean}
\end{align}
Consequently, we note that as \( N \to \infty \), \( \bar{S} \) approaches 0.5.

Similarly, substituting $\alpha=0$, $\Omega=1$ into equation \ref{eq:stdev} gives
\begin{align}
    \gamma & = 
    \sqrt{\frac{1}{N}\left(
    \frac{N-1}{2}
    (\bar{S} - 0)^2
    +
    \frac{N+1}{2}
    (\bar{S} - 1)^2
    \right)}
    \\ & = 
    \sqrt{\frac{1}{N}\left(
    \frac{N-1}{2}
    \bar{S}^2
    +
    \frac{N+1}{2}
    (\bar{S}^2 - 2\bar{S} +1)
    \right)}
    \\ & = 
    \sqrt{\left(
    \frac{1-\frac{1}{N}}{2}
    \bar{S}^2
    +
    \frac{1+\frac{1}{N}}{2}
    (\bar{S}^2 - 2\bar{S} +1)
    \right)}
    \\ & = 
    \sqrt{\left(
    \frac{\bar{S}^2}{2} - \frac{\bar{S}^2}{2N}\right)
    +
    \left(
    \frac{1}{2}+\frac{1}{2N}\right)
    (\bar{S}^2 - 2\bar{S} +1)
    }
    \\ & = 
    \sqrt{\left(
    \frac{\bar{S}^2}{2} - \frac{\bar{S}^2}{2N}\right)
    +
    \left(\frac{\bar{S}^2}{2} - \bar{S} +\frac{1}{2}\right)
    +
    \left(\frac{\bar{S}^2}{2N} - \frac{\bar{S}}{N} + \frac{1}{2N}\right)
    }
    \\ & = 
    \sqrt{\bar{S}^2\left(
    \frac{1}{2} - \frac{1}{2N} + \frac{1}{2} + \frac{1}{2N}\right)
    -
    \bar{S}\left(1 + \frac{1}{N}\right)
    +
    \left(\frac{1}{2} + \frac{1}{2N}\right)
    }
    \\ & = 
    \sqrt{\bar{S}^2
    -
    \bar{S} - \frac{\bar{S}}{N}
    +
    \frac{1}{2} + \frac{1}{2N}
    }
    % \\ & = 
    % \sqrt{\frac{1}{N}\left(
    % \frac{N-1}{2}
    % \left(\frac{1}{2} + \frac{1}{2N}\right)^2
    % +
    % \frac{N+1}{2}
    % \left(\frac{1}{2} + \frac{1}{2N} - 1\right)^2
    % \right)}
    \label{eq:stdev2}
\end{align}

Finally, substituting equation \ref{eq:mean} into equation \ref{eq:stdev2} we can fully simplify:
\begin{align}
    \gamma & =  
    \sqrt{\left(\frac{1}{2} + \frac{1}{2N}\right)^2
    -
    \left(\frac{1}{2} + \frac{1}{2N}\right) - \frac{\left(\frac{1}{2} + \frac{1}{2N}\right)}{N}
    +
    \frac{1}{2} + \frac{1}{2N}
    }
    \\ & = 
    \sqrt{
    \left(\frac{1}{4} + \frac{1}{4N^2} + \frac{1}{2N}\right)
    -
    \left(\frac{1}{2} + \frac{1}{2N}\right) - \left(\frac{1}{2N} + \frac{1}{2N^2}\right)
    +
    \frac{1}{2} + \frac{1}{2N}
    }
    \\ & = 
    \sqrt{
    \frac{1}{N^2}
    \left(\frac{1}{4} - \frac{1}{2}\right)
    +
    \frac{1}{4}
    }
    \\ & = 
    \sqrt{
    \frac{1}{4}
    \left(
    1 - \frac{1}{N^2}
    \right)
    }
    % \sqrt{\frac{1}{N}\left(
    % \frac{N-1}{2}
    % \left(\frac{1}{2} + \frac{1}{2N}\right)^2
    % +
    % \frac{N+1}{2}
    % \left(\frac{1}{2} + \frac{1}{2N} - 1\right)^2
    % \right)}
    \label{eq:stdev3}
\end{align}

\begin{figure*}[t!]
\centering
\includegraphics[width=0.6\textwidth]{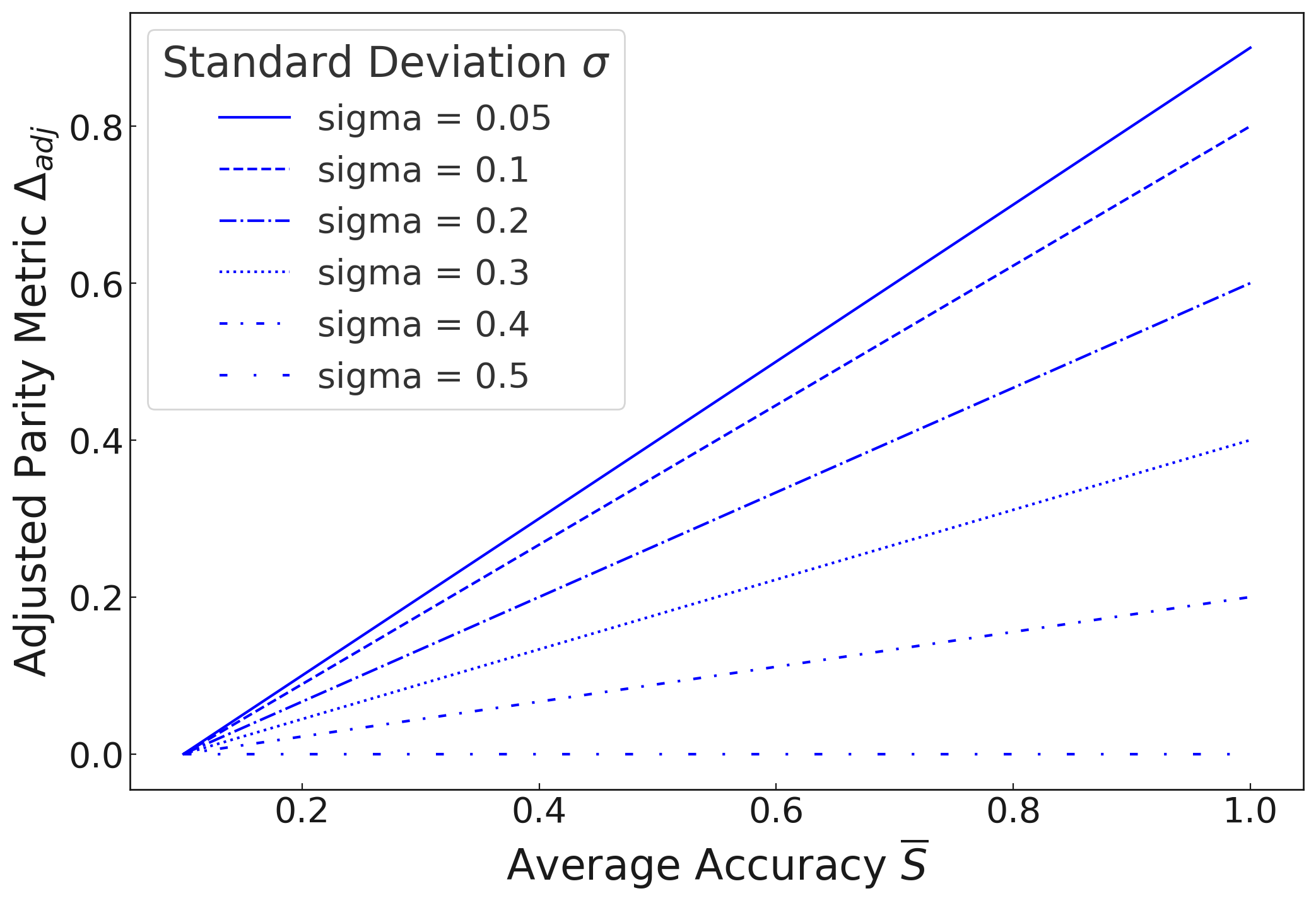}
\caption{Adjusted Parity Metric $\Delta_{\text{adj}}$ as a function of the Average Accuracy $\overline{S}$ for various values of the Standard Deviation $\sigma$ across domains. The baseline accuracy of a random predictor $S^R$ is set to 0.1, and the maximum standard deviation $\gamma$ is fixed at 0.5. The metric demonstrates how increasing performance inconsistency (higher $\sigma$) across domains reduces $\Delta_{\text{adj}}$, even when average accuracy is high.}
\label{fig6}
\end{figure*}

\section{Hyperparameter sensitivity on the COMPAS dataset}
Figure \ref{fig4b} shows the sensitivity of the DAP system to the $\beta$ and $\Omega$ hyperparameters on the COMPAS dataset. The results do not show the same trend witnessed on the Adult dataset with varying $\beta$ and $\Omega$. Increasing$\beta$ for a given $\Omega$ does not lower adjusted parity and improve fairness metrics as observed on the Adult dataset. The full set of results used to obtain graphs here and in the anonymous submission are placed with the zip file containing the code.

\begin{figure*}[t!]
\centering
\includegraphics[width=0.47\textwidth]{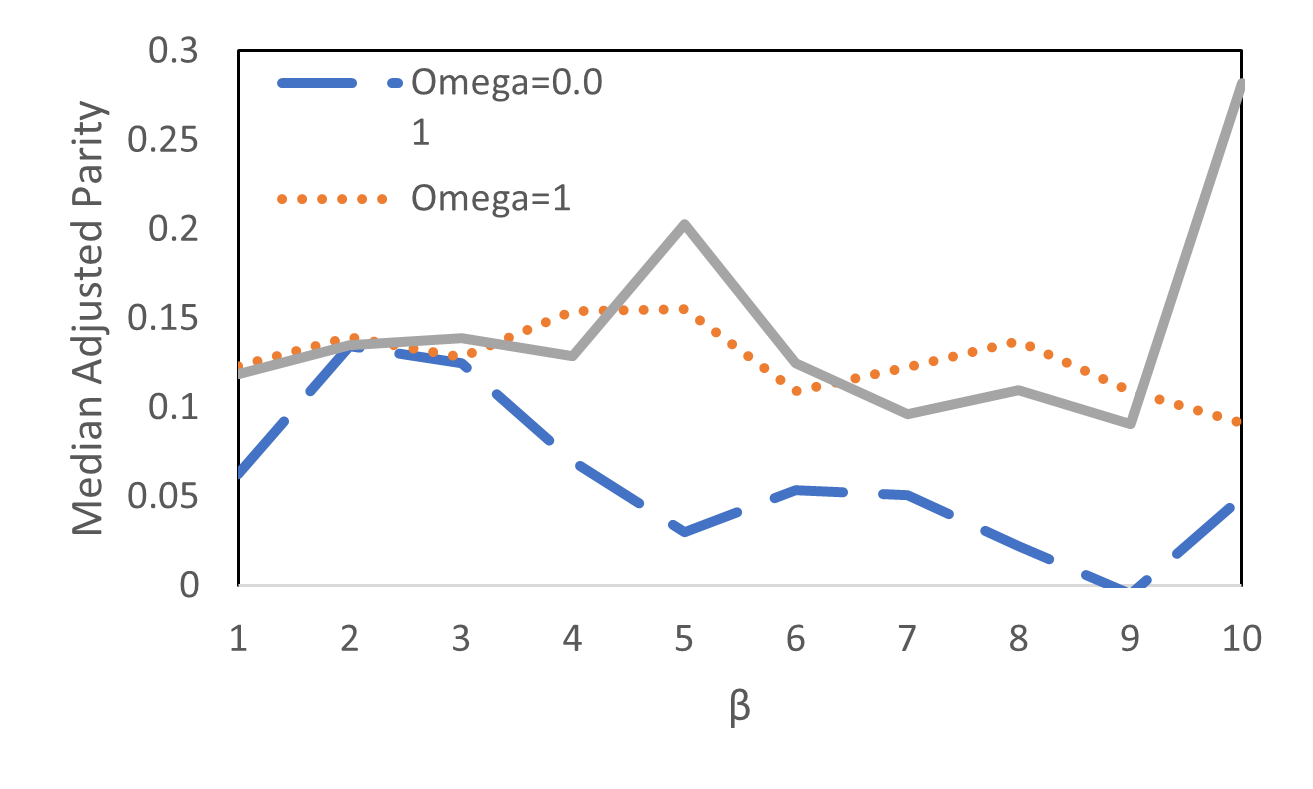}
\includegraphics[width=0.47\textwidth]{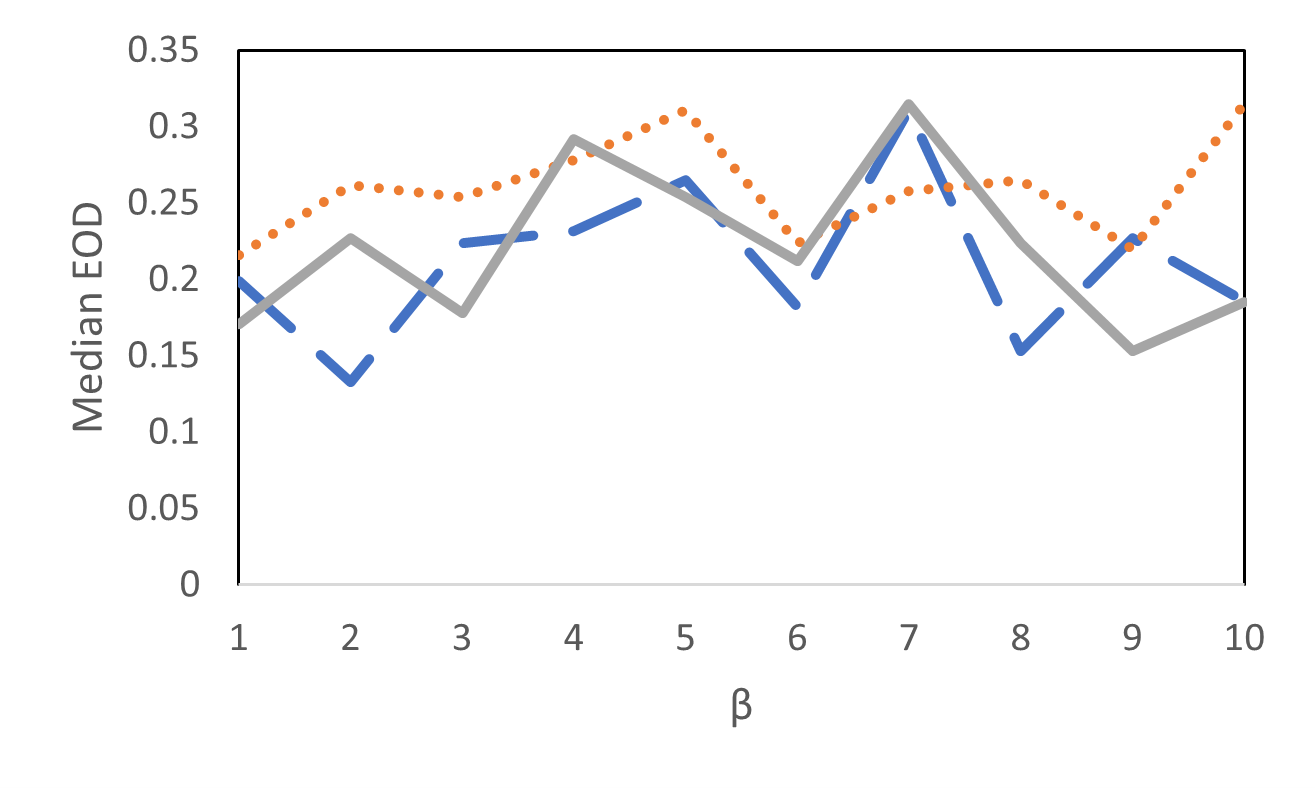} \\
\includegraphics[width=0.47\textwidth]{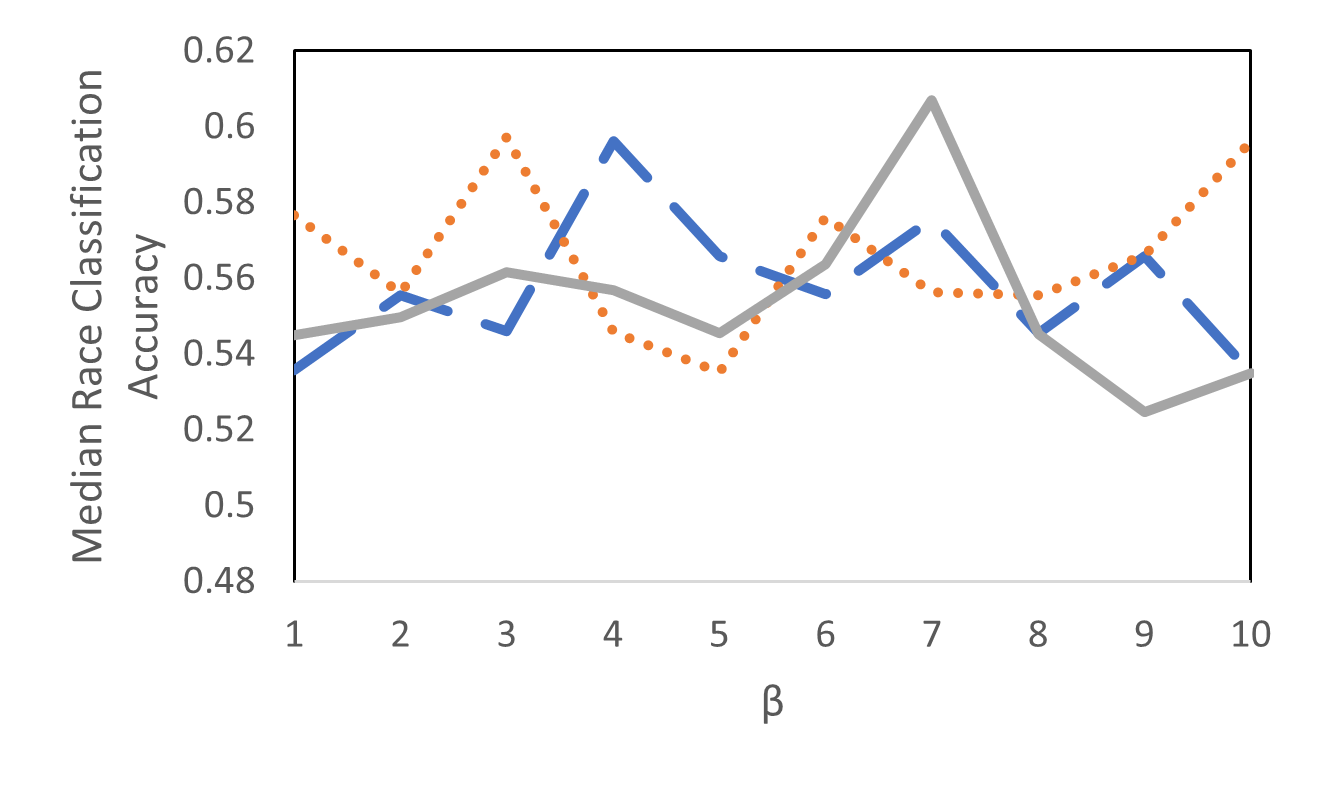}
\includegraphics[width=0.47\textwidth]{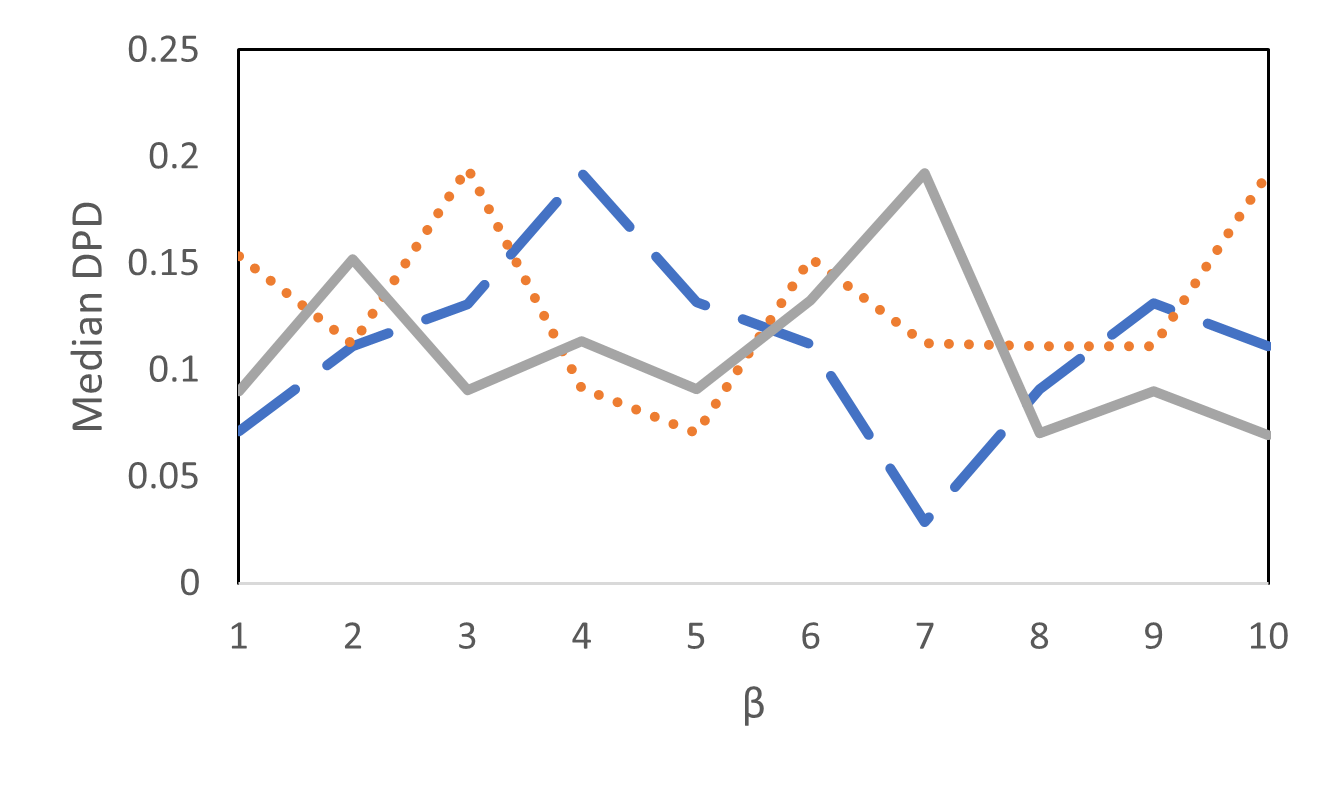}
\caption{Effect of altering $\Omega$ and $\beta$ on adjusted parity (top-left), EOD (top-right), gender accuracy (bottom-left), and DPD (bottom-right) on COMPAS dataset. Higher adjusted parity and lower EOD, DPD and gender accuracy are favourable.}
\label{fig7}
\end{figure*}

\end{document}